\documentclass{article}
\pdfoutput=1

\usepackage[nonatbib,final]{neurips_2018}

\usepackage[utf8]{inputenc} % allow utf-8 input
\usepackage[T1]{fontenc}    % use 8-bit T1 fonts
\usepackage{hyperref}       % hyperlinks
\usepackage{url}            % simple URL typesetting
\usepackage{booktabs}       % professional-quality tables
\usepackage{amsfonts}       % blackboard math symbols
\usepackage{nicefrac}       % compact symbols for 1/2, etc.
\usepackage{microtype}      % microtypography
\usepackage{graphicx,latexsym, amsmath}
\usepackage{color}
\usepackage{algorithm}

\usepackage[noend]{algpseudocode}
\usepackage{enumitem}
\usepackage{wrapfig}
\usepackage{atbegshi}% http://ctan.org/pkg/atbegshi
\usepackage{afterpage,lipsum} %,natbib

\usepackage{amsthm}
\usepackage{amsmath}
\usepackage{mathabx}
\usepackage{xspace}
\usepackage{color}
\usepackage{pgfplots, pgfplotstable}
\usepackage[colorinlistoftodos]{todonotes} %todo

\newcommand{\mn}[1]{\textcolor{blue}{{(MN: #1)}}}
\newcommand{\nl}[1]{\textcolor{purple}{{(NL: #1)}}}

\newcommand{\jb}[1]{\textcolor{brown}{{(JB: #1)}}}
\newcommand{\comment}[1]{}

\newcommand{\argmax}{\operatornamewithlimits{argmax}}

\def\etal{{\em et al.\xspace}}
\def\mapo{Memory Augmented Policy Optimization\xspace}
\def\MAPO{MAPO\xspace}

\def\REINFORCE{REINFORCE\xspace}

\renewcommand{\vec}[1]{\boldsymbol{\mathbf{#1}}}

\def\A{\mathcal{A}}
\def\piB{\pi_\B}
\def\piBo{\pi_\B^{\text{old}}}
\def\B{\mathcal{B}}
\def\Ba{\mathcal{B}_a}

\def\a{a}
\def\r{R}
\def\ba{\vec{a}}

\def\bx{\vec{x}}
\def\by{\vec{y}}
\def\ba{\vec{a}}
\def\bas{\vec{a}^*}

\def\objmml{\mathcal{O}_{\mathrm{MML}}}

\def\objmix{\mathcal{O}_{\mathrm{AUG}}}

\def\objer{\mathcal{O}_{\mathrm{ER}}}

\def\int{\mathrm{int}}

\DeclareMathOperator{\Var}{\mathrm{Var}}

\def\eg{{\em e.g.}}
\def\ie{{\em i.e.}}
\def\vs{{\em v.s.}\xspace}

\newcommand{\tabref}[1]{Table~\ref{#1}}
\newcommand{\figref}[1]{Figure~\ref{#1}}
\newcommand{\secref}[1]{Section~\ref{#1}}

\def\pit{\pi_\theta}
\def\expected{\mathbb{E}}
\def\oldpi{\pi_\theta^{old}}
\def\buf{\mathcal{B}}

\title{
{\mapo} for Program Synthesis and Semantic Parsing
}

\author{
  Chen Liang\\
  Google Brain \\
  \hspace{.4cm}\texttt{crazydonkey200@gmail.com}\hspace{.4cm}
  \And 
  Mohammad Norouzi \\
  Google Brain \\
  \hspace{.4cm}\texttt{mnorouzi@google.com}\hspace{.4cm} \\
  \And
  Jonathan Berant \\ 
  Tel-Aviv University, AI2 \\
  \texttt{joberant@cs.tau.ac.il} \\
  \And 
   Quoc Le \\
   Google Brain \\
  \texttt{qvl@google.com} \\
   \And 
   Ni Lao \\
   SayMosaic Inc. \\
   \texttt{ni.lao@mosaix.ai} \\
}

\begin{document}

\maketitle

\begin{abstract}
\vspace*{-.2cm}
We present \mapo (\MAPO), a simple and novel way to
leverage a memory buffer of promising trajectories
to reduce the variance of policy gradient estimates.
\MAPO is applicable to deterministic
environments with discrete actions, such as structured prediction and combinatorial optimization.
Our key idea is to express the expected return objective as a weighted sum of two terms: an
expectation over the high-reward trajectories inside a memory buffer, and a
separate expectation over trajectories outside of the buffer.
To design an efficient algorithm based on this idea, we propose:
(1) memory weight clipping to accelerate and stabilize training; 
(2) systematic exploration to discover high-reward trajectories;
(3) distributed sampling from inside and outside of the memory buffer to speed up training.
\MAPO improves the sample efficiency and robustness of policy gradient, especially on tasks with sparse rewards. 
We evaluate \MAPO on {\em weakly supervised} program synthesis from {\em natural language} (semantic parsing).
On the \textsc{WikiTableQuestions} benchmark, we improve the
state-of-the-art by {$2.6\%$}, achieving an accuracy
of {$46.3\%$}. On the \textsc{WikiSQL} benchmark, \MAPO achieves an accuracy of {$74.9\%$} with only weak supervision, outperforming several strong baselines with full supervision. Our source code is available at \href{https://goo.gl/TXBp4e}{{goo.gl/TXBp4e}}.

%% achives competitive results to other
%% state-of-the-art models trained with strong supervision.  In the
%% ablation study, we further show that {\MAPO} outperforms several
%% common PG baselines both in terms of sample efficiency and robustness.

%Within this formulation, we propose two techniques to (1) reduce the variance of the gradient estimation; (2) cold start and stabilize training.

% We separately study the problems of {\em generalization} and {\em exploration} in program
% synthesis and demonstrate that not only entropy regularized expected
% reward is a suitable objective for improving exploration, but also it
% helps finding the programs that generalize well to a heldout
% dataset. \todo{results}

%\nl{It is better to clearly define the problem as sample efficiency and robustness. } \mn{done}\\
% \nl{I don't think the training objective has anything to do with exploration.}

\comment{
  Deep reinforcement learning, especially policy gradient
  methods, has been successfully applied to many domains. Policy
  gradient methods often suffer from poor sample efficiency and
  robustness, especially given sparse rewards. In this work, we
  propose a new policy gradient objective, which reduces the variance
  in gradient estimation through stratified sampling from positive and
  negative trajectories, and speeds up and stabilizes training through
  a marginal likelihood constraint. We show that these two techniques
  allows policy gradient to significantly outperform previous. The
  proposed techniques are justified by both theoretical analysis and
  empirical ablation study.
}

\vspace*{-.2cm}
\end{abstract}

\comment{
\jb{Is having a test set considered zero-shot in program synthesis? Generalizing to the same task sampled from the same distribution is not zero-shot in my head.}
\nl{I think so :) The RL community has not done much about generalizing between different tasks/environments.}
\mn{I am against using zero-shot for this task. We can just call it generalization. I know people have used zero shot in this context but that is confusing and results from their weird mindset in which supervised learning is very different from RL}
\nl{sure let's bring Generalization back to title.}
}

\vspace*{-.1cm}
\section{Introduction}
\vspace*{-.2cm}
There has been a recent surge of interest in applying policy gradient
methods to various application domains including
program synthesis~\cite{liang2017nsm,guu2017language,zhong2017seq2sql,bunel2018leveraging},
dialogue generation~\cite{li2016deep,dasdialog2017},
architecture search~\cite{zoph2016,zoph2017}, game~\cite{silver2017mastering,mnih2016asynchronous}
and continuous control~\cite{peters2006,trpo2015}.
Simple policy gradient methods like
\REINFORCE~\cite{Williams92simplestatistical} use Monte Carlo samples
from the current policy to perform an {\em on-policy} optimization of
the expected return objective. This often leads to unstable learning dynamics and poor
sample efficiency, sometimes even underperforming random search~\cite{rs2018}.

The difficulty of gradient based policy optimization stems from a few
sources: \mbox{(1)}~policy gradient estimates have a large {\em
  variance}; \mbox{(2)}~samples from a randomly initialized policy
often attain small rewards, resulting in a slow training progress in
the initial phase (cold start); \mbox{(3)}~random policy samples do not explore the
search space efficiently and systematically. 
%because many samples can be repeated.
These issues can be especially prohibitive in applications such as
program synthesis and robotics~\cite{andrychowicz2017hindsight} where the search space is large
and the rewards are delayed and sparse. 
%\comment{
In such domains, a high reward is only achieved
after a long sequence of {\em correct} actions. For instance, in
program synthesis, only a few programs in the large combinatorial space of programs 
% implements the desired fun  
may correspond to the correct functional form. 
%}
\comment{
%poor trajectories usually doesn't have any gradients given sparse reward.
Policy gradient techniques have a hard
time separating the gradient of a high reward trajectory from the
noisy gradients of many poor trajectories
}  
Unfortunately, relying on policy samples to explore the space often
leads to forgetting a high reward trajectory unless it is re-sampled
frequently~\cite{liang2017nsm,pqt2018}.

\begin{figure}
%\vspace{-0.2in}
% \vspace{-0.5in}
  %\centering
  %\begin{minipage}{2.5in}
\hspace{-0.1in}
\begin{tabular}{cc}
\includegraphics[width=2.8in]{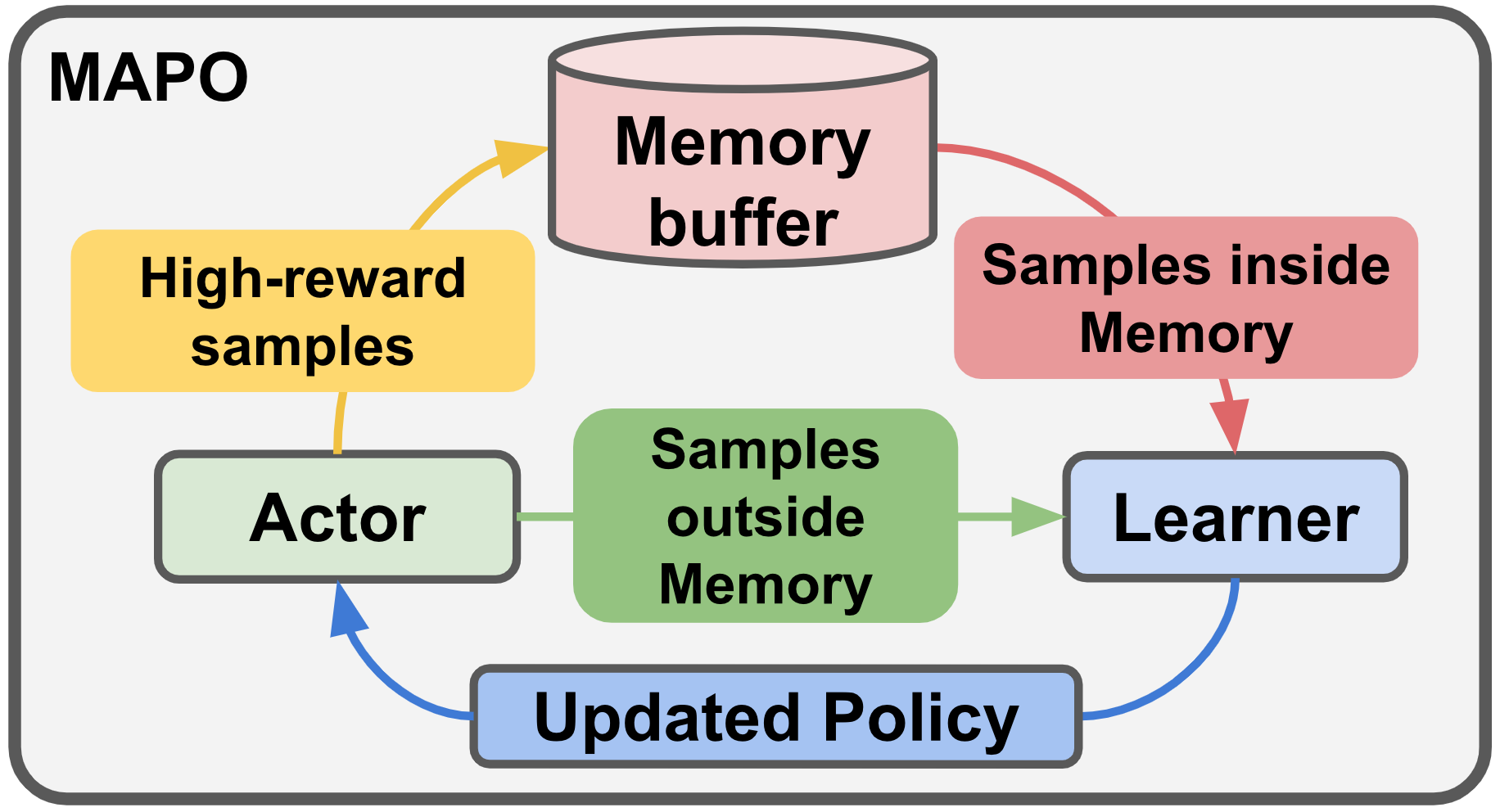}&
\hspace{-0.2in}
\includegraphics[width=2.8in]{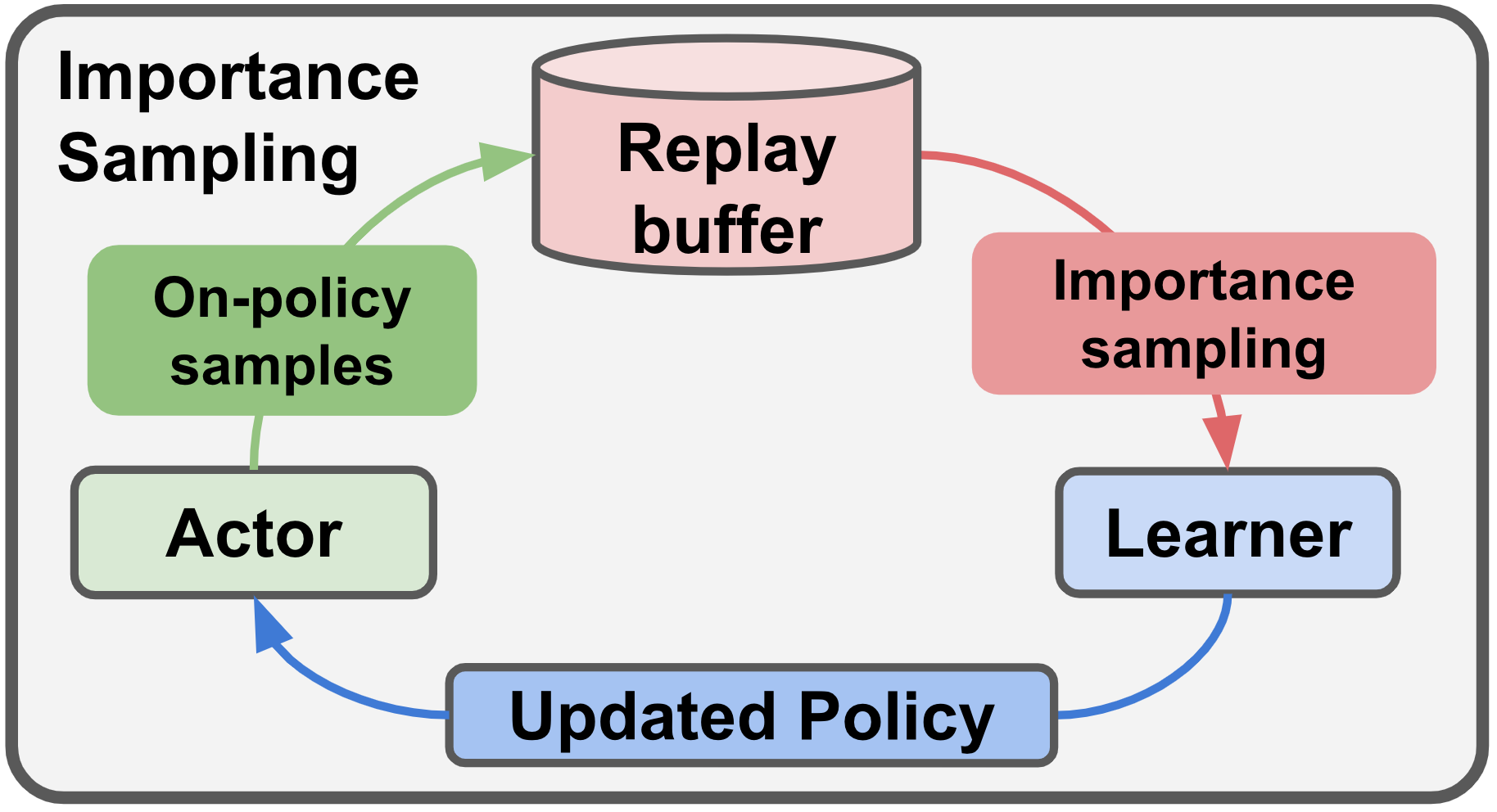}
\end{tabular}
 % \end{minipage}
 \vspace{-0.1in}
  \caption{Overview of \MAPO compared with experience replay using importance sampling. 
    }
  \label{fig:mapo-overview}
\vspace{-0.1in}
\end{figure}

Learning through reflection on past experiences (``experience replay'') is a promising direction to improve  data efficiency and learning stability. 
It has recently been widely adopted in various deep RL algorithms, but its theoretical analysis and empirical comparison are still lacking. 
%For example, studies of various RL tasks (e.g., Atari games, Mujoco tasks, physical robot arm, and DeepMind control suite) have all be expanding their replay buffer size to $10^6$. 
%However, recent study~\cite{zhang2017deeper} showcases that ``both a small replay buffer and a large replay buffer can heavily hurt the learning process''. 
As a result, defining the optimal strategy for prioritizing and sampling from past experiences remain an open question.
There has been various attempts to incorporate off-policy samples
within the policy gradient framework to improve the sample efficiency
of the REINFORCE and actor-critic algorithms
(\eg, ~\cite{degris2012,wang2016sample,ppo2017,espeholt2018impala}). Most
of these approaches utilize samples from an old policy through (truncated) importance sampling to obtain a low variance, but {\em biased}
estimate of the gradients. 
% To reduce the variance, one often truncates the importance weights if they become large. 
Previous work has aimed
to incorporate a replay buffer into policy gradient in the general RL
setting of stochastic dynamics and possibly continuous actions. 
By contrast, we focus on deterministic environments with discrete actions
and develop an {\em unbiased} policy
gradient estimator with low variance %. A comparison is shown in 
(Figure \ref{fig:mapo-overview}).
%This estimate does not require importance sampling based on an old version of the policy as the proposal. \nl{not sure about this}

%\nl{Exploration and fast convergence are two competing goals.}
%\nl{we actually don't need this assumption after clipping the replay buffer probability? I believe the only assumption needed is that the replay buffer has higher expected reward.}

% WChang
%just add 1 sentence after that, be more explicit to the reader about the problem w/ policy gradient methods u can forget
%i think it would add a bit more colour to the paper in the selling aspect
%here's a more general question , once the model is converged , do u think the memory buffer is still needed? do u think switching to regular policy gradient would probably result in same perf
%i.e., does the memory replay buffer solve the "learning without forgetting" problem

This paper presents \MAPO: a simple and novel %formulation of policy optimization
%for deterministic environments with discrete actions, which
way to incorporate a memory buffer of promising trajectories within
the policy gradient framework. 
We express the expected return objective as a weighted sum of an expectation over the trajectories inside the memory buffer
and a separate expectation over unknown trajectories outside of the buffer. 
The gradient estimates are unbiased and
attain lower variance. %provided that trajectories in the memory buffer have non-negligible probability.
%higher expected reward under the current policy.
% To alleviate the {\em cold-start} training issue, we perform random
% exploration to initialize the memory buffer with high reward
% trajectories. 
Because high-reward trajectories remain in the memory,
it is not possible to forget them.
To make an efficient algorithm for \MAPO, we propose 3 techniques: (1) memory weight clipping to accelerate and stabilize training; (2) systematic exploration of the search space to efficiently discover the high-reward trajectories; (3) distributed sampling from inside and outside of the memory buffer to scale up training; 

We assess the effectiveness of \MAPO on {\em weakly supervised} 
program synthesis from {\em natural language} (see
\secref{sec:weakps}). Program synthesis presents a unique opportunity
to study {\em generalization} in the context of policy optimization,
besides being an important real world application. On the challenging
\textsc{WikiTableQuestions}~\cite{pasupat2015tables} benchmark, \MAPO
achieves an accuracy of {$46.3\%$} on the test set,
significantly outperforming the previous state-of-the-art of
$43.7\%$~\cite{zhang2017macro}.  Interestingly, on the
\textsc{WikiSQL}~\cite{zhong2017seq2sql} benchmark, \MAPO achieves an
accuracy of {$74.9\%$} without the supervision of gold programs,
outperforming several strong {\em fully supervised} baselines. 
%jb: I think this sentence is not really necessary in the intro only in conclusions.
%We plan to apply the \MAPO framework to various other sequential decision making problems in the future.

\comment{
We propose three (independent) techniques to improve the efficiency
and robustness of policy gradient: 
(1) Stratified sampling
for reducing the variance in gradient estimation; (2) A constraint on the
marginal likelihood of high-reward trajectories to augment the policy
gradient objective, which alleviates ``cold-start training" while introducing little bias; (3) Systematic and efficient exploration of the search space in structured prediction or sequence generation tasks.
\comment{
(1) use stratified sampling
to reduce variance in gradient estimation; (2) use a constraint on the
marginal likelihood of high-reward trajectories to augment the policy
gradient objective to help cold start training without introducing
much bias; (3) use systematic exploration to efficiently explore the
search space in structured prediction or sequence generation tasks.}
}

\comment{
\mn{let's avoid REINFORCE instead of Reinforce. I think it is harder to read and in the title we don't want to use REINFORCE}
\nl{It looks like REINFORCE is used everywhere in literature, but we can use whatever we want for our own algorithm.}
\nl{Maybe we can use policy gradient in most places, and use REINFORCE only when we talk about the specific algorithm.}\jb{Policy gradient seems good to me.}
}

\vspace*{-.1cm}
\section{The Problem of Weakly Supervised Contextual Program Synthesis}
\label{sec:weakps}
\vspace*{-.2cm}

\begin{wraptable}{rh}{2.2in}
\setlength{\tabcolsep}{3pt}
%\begin{table}[ht] 
\vspace{-0.1in}
\centering 
%\scriptsize %
\small
  \begin{tabular}{ |l|l|c|r|r|}
  	\hline
    Year&Venue&Position&Event&Time\\
      	\hline
2001&Hungary&2nd&400m&47.12 \\
2003&Finland&1st&400m&46.69 \\
2005&Germany&11th&400m&46.62\\
2007&Thailand&1st&relay&182.05\\
2008&China&7th&relay&180.32 \\
     \hline
  \end{tabular}
\caption{%An example table and question-answer pair. 
\textbf{x}:  Where did the last 1st place finish occur? %\mn{great. better Q?}
\textbf{y}:  Thailand} 
\label{example-qa}
\vspace{-0.1in}
%\end{table}
\end{wraptable}

Consider the problem of learning to map a natural language question
$\bx$ to a structured query $\ba$ in a programming language such as
SQL (\eg, \cite{zhong2017seq2sql}), or %. Another similar problem is 
converting a textual
problem description into a piece of source code%, \eg, in python, 
as in
programming competitions~(\eg, ~\cite{balog2017deepcoder}). We call these
problems {\em contextual program synthesis} and 
aim at tackling them in a weakly supervised setting -- i.e.,
no correct action sequence $\ba$, which corresponds to a gold program, is given as part of the training
data, and training needs to solve the hard problem of exploring a
large program space. Table \ref{example-qa} shows an example question-answer pair. The model needs to first discover the programs that can generate the correct answer in a given context, and then learn to generalize to new contexts.

We formulate the problem of {\em weakly supervised} contextual program
synthesis as follows: %Suppose we are Given a context $\bx$, one needs
to generate a program by using a parametric function,
$\hat{\ba} = f(\bx; \theta)$, where $\theta$ denotes the model
parameters.  The quality of a program $\hat{\ba}$ is
measured by a scoring or {\em reward} function
$R(\hat{\ba} \mid \bx, \by)$. The reward function may evaluate a program by executing it on a real environment and comparing the output against the correct answer. 
For example, it is natural to define a binary reward that is 1 when the output equals the answer and 0 otherwise.
We assume that the context $\bx$
includes both a natural language input and an environment, for example an interpreter or a database, on which the
program will be executed. Given a dataset of context-answer pairs,
$\{(\bx_i,
\by_i)\}_{i=1}^N$, the goal is to find optimal parameters $\theta^*$
that parameterize a mapping of $\bx \to \ba$ with maximum empirical
return on a {\em heldout test set}. 

%% Since no RL training is performed
%% at test time, we also call the problem {\em zero-shot program
%%   synthesis}.

% \comment{Hence, we are particularly interested in
% {\em generalization} in the context of reinforcement learning.}
%\jb{the reward also depends on the environment for which there is no
%notation if i understand correctly.}  \mn{I am assuming the
%environment is in $\bx$ too. We should clarify it later.}

One can think of the problem of contextual program synthesis as an
instance of {\em reinforcement learning (RL)} with {\em sparse
terminal rewards} and {\em deterministic transitions}, for which {\em
generalization} plays a key role. There has been some recent attempts
in the RL community to study generalization to unseen initial
conditions (\eg~\cite{rajeswaran2017towards,gotta2018gotta}). However,
most prior work aims to maximize empirical return on the training
environment~\cite{arcade2013,gym2016}. The problem of contextual
program synthesis presents a natural application of RL for which
generalization is the main concern.
% Accordingly, we separate the effect of generalization from
% exploration in our experiments and study the effect of each problem
% in isolation.

% \comment{ \jb{General comment: intro is written for RL reviewers not
%     semantic parsing reviewers (from a SP perspective for example
%     almost all of 2 is the standard setup and requires less
%     justification), which is fine just need to be aware and direct the
%     submissions towards the correct reviewers.} \mn{I moved it to
%     section 2. I think repeating the problem statement does not hurt}
% }

\vspace*{-.1cm}
\section{Optimization of Expected Return via Policy Gradients}
\vspace*{-.2cm}
To learn a mapping of (context $\bx$) $\to$ (program $\ba$), we
optimize the parameters of a conditional distribution
$\pit(\ba \mid \bx)$ that assigns a probability to each program given
the context. That is, $\pit$ is a distribution over the {\em
countable} set of all possible programs, denoted $\mathcal{A}$. Thus
$\forall \ba \in \mathcal{A}:~ \pit(\ba
\mid \bx) \geq 0$ and $\sum_{\ba \in \mathcal{A}} \pit(\ba \mid \bx) =
1$. Then, to synthesize a program for a novel context, one finds the
most likely program under the distribution $\pit$ via exact or
approximate inference
%\begin{equation}
$\hat{\ba} ~\approx~ \argmax_{\ba \in \mathcal{A}} \pit(\ba \mid \bx)~.$
%\end{equation}

{\em Autoregressive} models present a tractable family of distributions 
%over programs 
that estimates the probability of a sequence of tokens, one
token at a time, often from left to right. To handle variable sequence
length, one includes a special {\em end-of-sequence} token at the end
of the sequences. We express the probability of a program $\ba$ given
$\bx$ as
$%\begin{equation}
\pit(\ba \mid \bx) ~\equiv~
\prod\nolimits_{i=t}^{\lvert \ba \rvert} \pit(\a_t \mid \ba_{<t},\bx)~,
$%\end{equation}
where $\ba_{<t} \equiv (\a_1, \ldots, \a_{t-1})$ denotes a prefix of
the program $\ba$. One often uses a recurrent neural network
(\eg~\cite{hochreiter1997long}) to predict the probability of each
token given the prefix and the context.

In the absence of ground truth programs, policy gradient techniques
present a way to optimize the parameters of a stochastic policy $\pit$
via optimization of {\em expected return}. Given a training dataset of
context-answer pairs, $\{(\bx_i, \by_i)\}_{i=1}^N$,
the objective is expressed as $\mathbb{E}_{\ba \sim \pit(\ba \mid
  \bx)}\:\r(\ba \mid \bx, \by)$.  The reward function $\r(\ba \mid
\bx, \by)$ evaluates a complete program $\ba$, based on the context
$\bx$ and the correct answer $\by$. 
These assumptions characterize the problem of program synthesis well, but they also apply to many other discrete optimization and structured prediction domains. 
% Note that these assumptions are required for the exact unbiased gradients, but we also expect the \MAPO algorithm to benefit many other discrete RL tasks. 
%on which these assumptions apply to a big extent.

{\bf Simplified notation.} In what follows, we simplify the notation by
dropping the dependence of the policy and the reward on $\bx$ and
$\by$. We use a notation of $\pit(\ba)$ instead of $\pit(\ba \mid
\bx)$ and $\r(\ba)$ instead of $\r(\ba \mid \bx, \by)$, to make the
formulation less cluttered, but the equations hold in the general
case.

We express the expected return objective in the simplified notation as,
\begin{eqnarray}
  %J_{\text{ER}}
  \label{eq:objer}
  \objer(\theta) ~=~ \sum_{\ba \in \mathcal{A}} \pit(\ba)\:\r(\ba) ~=~ \mathbb{E}_{\ba \sim \pit(\ba)}\:\r(\ba)~.
\end{eqnarray}
The \REINFORCE~\cite{Williams92simplestatistical} algorithm presents
an elegant and convenient way to estimate the gradient of the expected
return \eqref{eq:objer} using Monte Carlo (MC) samples. Using $K$
trajectories sampled {\em i.i.d.} from the current policy $\pit$,
denoted $\{\ba^{(1)},\ldots,\ba^{(K)}\}$, the gradient estimate can be
expressed as,
\begin{eqnarray}
  \nabla_\theta \objer(\theta) = \expected_{\ba \sim \pit(\ba)}\:\nabla\log \pit(\ba)\:\r(\ba)
  ~\approx~ \frac{1}{K} \sum_{k=1}^K \nabla\log \pit(\ba^{(k)})\:[\r(\ba^{(k)})-b]~,
  \label{eq:pg}
\end{eqnarray}
where a baseline $b$ is subtracted from the returns to reduce the
variance of gradient estimates. 
% A close-to-optimal form of a baseline is the on-policy average of the returns $\mathbb{E}_{\ba \sim \pit(\ba)}\:\r(\ba)$, often approximated empirically as $b ={\sum_k\r(\ba^{(k)})}/{K}$. 
This formulation enables direct
optimization of $\objer$ via MC sampling from an unknown search
space, which also serves the purpose of exploration. To improve such
exploration behavior, one often includes the entropy of the policy as
an additional term inside the objective to prevent early
convergence. However, the key limitation of the formulation stems from
the difficulty of estimating the gradients accurately only using a few
{\em fresh} samples.

\vspace*{-.1cm}
\section{\MAPO: \mapo}
\vspace*{-.2cm}
We consider an RL environment with a finite number of discrete
actions, deterministic transitions, and deterministic terminal returns.
In other words, the set of all possible action trajectories $\A$ is
countable, even though possibly infinite, and re-evaluating the return
of a trajectory $R(\ba)$ twice results in the same value. These assumptions
characterize the problem of program synthesis well, but also apply to many structured prediction problems \cite{ross2011reduction,nowozin2011structured} and
combinatorial optimization domains (\eg, ~\cite{bello2016neural}).

\comment{ For simplicity we only consider terminal returns and not intermediate
rewards even though they can be incorporated.  }

% Our goal is to optimize the expected return objective \eqref{eq:objer} via gradient ascent. 
To reduce the variance in gradient estimation and prevent forgetting high-reward trajectories, we introduce a memory buffer, which saves a set of promising trajectories denoted $\B \equiv \{(\ba^{(i)})\}_{i=1}^M$. Previous works~\cite{liang2017nsm,abolafia2018neural,wu2016gnmt} utilized a memory buffer by adopting a training objective similar to
\begin{equation}
\label{eqn:ml-aux}
\objmix(\theta) = \lambda  \objer(\theta) + (1 - \lambda) \sum_{\ba \in \mathcal{B}} \log\pit(\ba),
\end{equation}
which combines the expected return objective with a maximum likelihood objective over the memory buffer $\mathcal{B}$. This training objective is not directly optimizing the expected return any more because the second term introduces bias into the gradient. When the trajectories in $\mathcal{B}$ are not gold trajectories but high-reward trajectories collected during exploration, uniformly maximizing the likelihood of each trajectory in $\mathcal{B}$ could be problematic. For example, in program synthesis, there can sometimes be spurious programs~\cite{pasupat2016inferring} that get the right answer, thus receiving high reward, for a wrong reason, e.g., using $2+2$ to answer the question ``what is two times two''. Maximizing the likelihood of those high-reward but spurious programs will bias the gradient during training.

We aim to utilize the memory buffer in a principled way. Our key insight is that one
can re-express the expected return objective as a weighted sum of two terms: 
an expectation over the trajectories inside the memory buffer, and a separate expectation over the trajectories outside the buffer,
\begin{align}
  \objer(\theta) ~=~ & \sum_{\ba \in \B} \pit(\ba)\:R(\ba)\!\!\!\!\!\!\!\!\!\!\!\!\!\!\!\!\!\!\!\!\!&+\:\:\:\:\:\: & \sum_{\ba \in (\A-\B)} \pit(\ba)\:R(\ba) & \\
    ~=~ & \piB \,\underbrace{\expected_{\ba \sim \pit^+(\ba)}\,R(\ba)}_{\text{Expectation inside}\:\B}\!\!\!\!\!\!\!\!\!\!\!\!\!\!\!\!\!\!&+\:\:\:\:\:\: &  (1 - \piB)\,\underbrace{\expected_{\ba \sim \pit^-(\ba)}\,R(\ba)}_{\text{Expectation outside}\:\B}~, &
  \label{eq:two-expectation-pg}
\end{align}
where $\A-\B$ denotes the set of trajectories not included in
the memory buffer, $\piB = \sum_{\ba \in \B} \pit(\ba)$ denote the total probability of the
trajectories in the buffer,
and $\pit^+(\ba)$ and $\pit^-(\ba)$ denote a normalized
probability distribution inside and outside of the buffer,
\begin{equation}
\pit^+(\ba) =
\begin{cases}
{\pit(\ba)}/{\piB}       & ~\text{if}~\ba \in \B\\
0                        & ~\text{if}~\ba \not\in \B\\
\end{cases},~~~~~~~\pit^-(\ba) =
\begin{cases}
0                        & ~\text{if}~\ba \in \B\\
{\pit(\ba)}/{(1 - \piB)} & ~\text{if}~\ba \not\in \B
\end{cases}~.
\label{eq:renorm-prob}
\end{equation}

% The key intuition of the \MAPO algorithm is to use enumeration to evaluate the former expectation and $\piB$, and MC sampling to compute the latter expectation.
% So the gradient will be estimated as
% Using $K$ trajectories
% $\{\ba^{(1)},\ldots,\ba^{(K)}\}$ sampled {\em i.i.d.} from current
% $\pit^-$, %\ie~the unexplored region of the space, 
The policy gradient can be expressed as,
\begin{align}
    \nabla_\theta \objer(\theta) = \piB \, \expected_{\ba \sim \pit^+(\ba)}\,\nabla\log \pit(\ba)\r(\ba) + (1 - \piB) \, \expected_{\ba \sim \pit^-(\ba)}\,\nabla\log \pit(\ba)\r(\ba).
\label{eq:two-grad}
\end{align}
% \begin{equation}
%   \nabla_\theta \objer(\theta) ~\approx~ \sum_{\ba \in \B} \nabla\pit(\ba)\:[\r(\ba)-b] \:+\: \frac{1 - \piB}{K} \sum_{k=1}^K \nabla\log \pit(\ba^{(k)})\:[\r(\ba^{(k)})-b]~.
%   \label{eq:two-expectation-pg}
% \end{equation}

The second expectation can be estimated by sampling from $\pit^-(\ba)$, which can be done through rejection sampling by sampling from $\pit(\ba)$ and rejecting the
sample if $\ba \in \B$. If the memory buffer only contains a small number of trajectories, the first expectation can be computed exactly by enumerating all the trajectories in the buffer. The variance in gradient estimation is reduced because we get an exact estimate of the first expectation 
while sampling from a smaller stochastic space of measure $(1 - \piB)$. If the memory buffer contains a large number of trajectories, the first expectation can be approximated by sampling. Then, we get a {\em stratified sampling} estimator of the gradient. The trajectories inside and outside the memory buffer are two mutually exclusive and collectively exhaustive strata, and the variance reduction still holds. 
%
% From \eqref{eq:two-expectation-pg} and \eqref{eq:two-grad}, we can see that both the expected return objective and its gradient can be decomposed into a weighted sum of two expectations over the trajectories inside and outside the memory buffer, and 
The weights for the first and second expectations are $\piB$ and $1-\piB$ respectively. We call $\piB$  the {\em memory weight}.% in the training objective.
%since it  decides how to distribute weight between the two expectations over trajectories inside and outside the memory buffer.

In the following we present $3$ techniques to make an efficient algorithm of \MAPO.

\vspace*{-.2cm}
\subsection{Memory Weight Clipping}
\vspace*{-.2cm}
\label{sec:mml}

Policy gradient methods usually suffer from a cold start problem. % and the training is often brittle. 
A key observation is that a ``bad'' policy, one that achieves low expected return, will assign small probabilities to the high-reward trajectories, which in turn causes them to be ignored during gradient estimation. So it is hard to improve from a random initialization, \ie, the cold start problem, or to recover from a bad update, \ie, the brittleness problem. 
%To overcome these problems,
Ideally we want to force the policy gradient estimates to pay at least some attention to the high-reward trajectories. %Specifically, 
Therefore, we adopt a clipping mechanism over the memory weight $\piB$, which ensures that the memory weight is greater or equal to $\alpha$, \ie~$, \piB \ge \alpha$, otherwise clips it to $\alpha$. So the new gradient estimate is, 
\begin{equation}
    \nabla_\theta \objer^c(\theta) = \piB^c \, \expected_{\ba \sim \pit^+(\ba)}\,\nabla\log \pit(\ba)\r(\ba) + (1 - \piB^c) \, \expected_{\ba \sim \pit^-(\ba)}\,\nabla\log \pit(\ba)\r(\ba),
\end{equation}
where $\piB^c = \max(\piB, \alpha)$ is the clipped memory weight. 
%effectively optimizing,
% \begin{equation} \label{eqn:mml}
% \objmml(\theta) = \frac{1}{N} \sum_i \log \sum_{\ba \in
%   \mathcal{B}_i} \pit(\ba) = \frac{1}{N} \log \prod_i \sum_{\ba \in
%   \mathcal{B}_i} \pit(\ba).
% \end{equation}
At the beginning of training, the clipping is active and introduce a bias, but accelerates and stabilizes training. 
Once the policy is off the ground, the memory weights are almost never clipped given that they are naturally larger than $\alpha$ and the gradients are not biased any more. 
See section \ref{sec:expt-clip} for an empirical analysis of the clipping.

\vspace*{-.2cm}
\subsection{Systematic Exploration}
\vspace*{-.2cm}
\label{sec:systematic}

\begin{wrapfigure}{r}{2in}
\begin{minipage}{2in}
\vspace{-.5in}
\begin{algorithm}[H] %[ht!]
\caption{Systematic Exploration}
\label{alg:rand-explore}
\begin{algorithmic}
\State \textbf{Input:} context $\bx$, policy $\pi$, 
%vocabulary $V$,
%end of sentence token \text{EOS}, 
%, a function to compute the accumulated rewards of a sequence $R(a_{1:t})$, a threshold $\beta$
%\State \textbf{Initialize:}
%exploration buffer of 
fully explored sub-sequences $\buf^{e}$, 
% \leftarrow \emptyset$,\\
%\State \textbf{Initialize:}
%memory buffer of 
high-reward sequences $\buf$ 
%, concatenation operation $\Vert$  % \leftarrow \emptyset$
%\State \textbf{Procedure:}
%\For{$i \text{\ in\ } 1...N$}
\State \textbf{Initialize:}  empty sequence $a_{0:0}$ % \leftarrow \{\}$
\While {true}
\State $V=%_{valid} =% V \cap 
\{a\ |\ a_{0:t-1} \Vert a \notin B^{e}\}$
\If {$V== \emptyset$}
\State $\buf^{e} \leftarrow \buf^{e} \cup a_{0:t-1}$
\State \textbf{break}
\EndIf
\State sample $a_t \sim \pi^V(a|a_{0:t-1})$
%\footnote{$\pi^V$ is $\pi$ renormalized over $V$.}%_{valid}$
\State $a_{0:t} \leftarrow a_{0:t-1} \Vert a_t$ 
\If {$a_t == \text{EOS}$}
  \If {$R(a_{0:t}) > 0$}  
      \State $\buf \leftarrow \buf \cup a_{0:t}$
  \EndIf
  \State $\buf^{e} \leftarrow \buf^{e} \cup a_{0:t}$
  \State \textbf{break}
\EndIf
\EndWhile
%\EndFor
%\State \textbf{Output:} $Z^{+}$
\label{alg-sys}
\end{algorithmic}
\end{algorithm}
% \vspace{-0.2in}
    \end{minipage}
  \end{wrapfigure}

%The two techniques introduced above assume the memory buffer contains high reward trajectories. 
To discover high-reward trajectories for the memory buffer $\mathcal{B}$, we need to efficiently explore the search space. Exploration using policy samples suffers from repeated samples, which is a waste of computation in deterministic environments. So we propose to use systematic exploration to improve the efficiency. More specifically we keep a set $\buf^{e}$ of fully explored
partial sequences, which can be efficiently implemented
using a bloom filter. 
Then, we use it to enforce a %random or optimized 
policy to only take actions that lead to unexplored sequences. Using a bloom filter we can store billions of
sequences in $\buf^{e}$ with only several gigabytes of memory. 
% explore new sequences as follows. At each step, when the policy needs
% to pick a token to output, we mask out the tokens that can only lead
% to sequences in $Z^{e}$. We then sample from the rest according to the
% probability assigned by the policy (renormalized among the valid
% tokens). When the end of a sequence is reached or when there is no
% valid tokens to output anymore, which means all the sequences that
% have the current sequences as prefix is already explored, we save the
% current sequence into $Z^{e}$. When the sequence results in high
% rewards, we save it in $Z^{+}$. 
The pseudo code of this approach is shown in Algorithm \ref{alg:rand-explore}.
We warm start the memory buffer using systematic exploration from random policy as it can be trivially parallelized. 
In parallel to training, we continue the systematic exploration with the current policy to discover new high reward trajectories.

\vspace*{-.1cm}
\subsection{Distributed Sampling} 
%via Stratified Sampling Stratified sampling for variance minimization}
\vspace*{-.1cm}

\begin{wrapfigure}{r}{2.5in}
      \begin{minipage}{2.5in}
%\afterpage{
\vspace{-0.5in}
\begin{algorithm}[H] %[ht!]
\caption{MAPO}\label{alg:ccpg}
\begin{algorithmic}
\State \textbf{Input:}  
data $\{(\bx_i, \by_i)\}_{i=1}^N$, 
%memory and exploration buffer 
memories $\{(\buf_i, \buf_i^e)\}_{i=1}^N$,
%thresholds
constants $\alpha$, $\epsilon$, $M$ %initial 
%\State \textbf{Initialize:}  $\pi_\theta$ randomly
%\State \textbf{Actor Repeat:}
\Repeat \Comment{\textbf{for all actors}}
%   \State $\oldpi \leftarrow \pit$
  \State Initialize training batch $D \leftarrow \emptyset$
  \State Get a batch of inputs $C$
  %\For {i in {1...K}}
  \For {$(\bx_i, \by_i, \buf^{e}_i,\buf_i) \in C$}
  %\State $\mathrm{Systematic Exploration}
  \State $\mathrm{Algorithm1}(\bx_i, \oldpi, \buf^{e}_i,\buf_i)$
  %Perform 1 iteration of systematic exploration according to Algorithm \ref{alg-sys} 
  % Draw $n_s$ systematic exploration samples using $\oldpi$ and add them to $\mathcal{B}_i$ according to \ref{alg-sys}
 % \State Sample $\ba_i^+$ from $\mathcal{B}_i$ according to $\oldpi$ 
  \State Sample $\ba_i^+ \sim \oldpi$ over $\buf_i$ 
  \State $w_i^+ \leftarrow \max(\oldpi(\mathcal{B}_i), \alpha)$
%   \If {use baseline}
%   \State $b_i=\oldpi(\mathcal{B}_i)$
%   \Else 
%   \State $b_i=0$
%   \EndIf
  \State $D \leftarrow D \cup (\ba_i^+, R(\ba_i^+), w_i^+)$
  \State Sample $\ba_i \sim \oldpi$ 
  \If {$\ba_i \notin \mathcal{B}_i$}
  \State $w_i \leftarrow (1 - w_i^+)$
  \State $D \leftarrow D \cup 
  (\ba_i, R(\ba_i), w_i)$ 
  \EndIf
  \EndFor
  \State Push $D$ to training queue
\Until{converge or early stop}
%\end{ALC@g}%\eindent
%\State \textbf{Learner Repeat:}
\Repeat \Comment{\textbf{for the learner}}
%\begin{ALC@g} %\bindent
  \State Get a batch $D$ from training queue
  \For { $(\ba_i, R(\ba_i), w_i) \in D$ }
    %\If {$A(\ba_i) > 0 \vee \pit(\ba_i) > \epsilon $}
%\Continue
  \State $\mathrm{d} \theta \leftarrow \mathrm{d}   \theta + w_i\:R(\ba_i)\:\nabla \log \pit(\ba_i)$
 %\EndIf
  \EndFor
  \State update $\theta$ using $\mathrm{d} \theta$
%\end{ALC@g} %\eindent
  \State $\oldpi \leftarrow \pit$  \Comment{once every M batches}
\Until{converge or early stop}
%a certain number of steps is reached}
\State \textbf{Output:} final parameters $\theta$
\end{algorithmic}
\end{algorithm}
\vspace{-0.3in}
%} % end of "afterpage" group
  \end{minipage}
  \end{wrapfigure}
%\lipsum[4-5]

An exact computation of the first expectation 
of \eqref{eq:two-expectation-pg} requires an enumeration over the memory buffer. The cost of gradient computation will grow linearly {\em w.r.t} the number of trajectories in the buffer, so it can be prohibitively slow when the buffer contains a large number of trajectories. 
%For example, in our experiments, we find that some contexts have up to $300$ trajectories in the memory buffer. 
%If $\lvert\buf\rvert$ is small, enumeration is possible at a negligible cost. 
%If $\lvert\buf\rvert$ is large, enumeration can become prohibitive.  
% One solution is to simply truncate the memory buffer to keep only a few trajectories with the highest probabilities.  
% However, this simple approach may lead to suboptimal result, because one often benefits from keeping all of the promising programs in the buffer to let model select the ones that generalize well. 
% For example, in program synthesis, there can sometimes be many programs that compute the correct answer for a certain question, but only a handful will generalize and others are spurious (getting the correct answer because of luck, e.g., using $2+2$ to answer the question ``what is two times two''). The generalizable programs could be discarded during the truncation of the memory buffer, making it hard for the model to overcome the spurious programs. 
Alternatively,  
we can approximate the first expectation using sampling.
As mentioned above, this can be viewed as {\em stratified sampling} and the variance is still reduced. 
Although the cost of gradient computation now grows linearly {\em w.r.t} the number of samples instead of the total number of trajectories in the buffer, the cost of sampling still grows linearly {\em w.r.t} the size of the memory buffer because we need to compute the probability of each trajectory with the current model. 

A key insight is that if the bottleneck is in sampling, the cost can be distributed through an actor-learner architecture similar to~\cite{espeholt2018impala}. 
See the Supplemental Material \ref{apdx:actor-learner} for a figure depicting the actor-learner architecture.
The actors each use its model to sample trajectories from 
inside the memory buffer through renormalization ($\pi_{\theta}^+$ in \eqref{eq:renorm-prob}), 
and uses rejection sampling to pick trajectories from outside the memory ($\pi_{\theta}^-$ in \eqref{eq:renorm-prob}). 
It also computes the weights for these trajectories using the model. 
These trajectories and their weights are then pushed to
a queue of samples. 
The learner fetches samples from the queue and uses
them to compute gradient estimates to update the parameters. 
By distributing the cost of sampling to a set of actors, the training can be accelerated almost linearly {\em w.r.t} the number of actors. In our experiments, we got a $\sim$20 times speedup from distributed sampling with 30 actors.

\vspace*{-.2cm}
\subsection{Final Algorithm}
\vspace*{-.2cm}
\label{sec:ccpg-alg}

The final training procedure is summarized in
Algorithm~\ref{alg:ccpg}. As mentioned above, we adopt the
actor-learner architecture for distributed training. It uses multiple
actors to collect training samples asynchronously and one learner for
updating the parameters based on the training samples. Each actor
interacts with a set of environments to generate new trajectories. 
%An environment includes an interpreter and a natural language utterance. \nl{redundant?}
For efficiency, an actor uses a stale policy ($\pit^{old}$),
which is often a few steps behind the policy of the learner and will
be synchronized periodically. To apply \MAPO, each actor also maintains a
memory buffer $\mathcal{B}_i$ to save the high-reward trajectories. To
prepare training samples for the learner, the actor picks $n_b$
samples from inside $\mathcal{B}_i$ and also performs rejection sampling with $n_o$ on-policy samples,
both according to the actor's policy $\pit^{old}$.
We then use the actor policy to compute a weight
$max(\pit(\mathcal{B}), \alpha)$ for the
samples in the memory buffer, and use $1 - max(\pit(\mathcal{B}), \alpha)$
for samples outside of the buffer. These samples are pushed to a queue and
the learner reads from the queue to compute gradients and update the parameters.

\vspace*{-.2cm}
\section{Experiments}
\vspace*{-.2cm}
\label{sec:experiment}

We evaluate MAPO on two program synthesis from natural language (also known as \textit{semantic parsing}) benchmarks, \textsc{WikiTableQuestions} and \textsc{WikiSQL}, which requires generating programs to query and process data from tables to answer natural language questions. We first compare MAPO to four common baselines, and ablate systematic exploration and memory weight clipping to show their utility. Then we compare MAPO to the state-of-the-art on these two benchmarks. On \textsc{WikiTableQuestions}, MAPO is the first RL-based approach that significantly outperforms the previous state-of-the-art. On \textsc{WikiSQL}, MAPO trained with weak supervision (question-answer pairs) outperforms several strong models trained with full supervision (question-program pairs). 

% Before diving into details, we define the task as follows:
% given an interpreter $\mathbb{I}$ that has built-in functions and databases and a natural language utterance $q=(w_1, w_2, ..., w_m)$ as input, produce a program $\ba$ in a domain specific language (DSL) that, when executed on $\mathbb{I}$, generates the right answer $\by$. 
% We use $\bx$ to denote the input, including the utterance and the interpreter. Since the true program $\ba_i$  
% %expertise to annotate and is tied to a specific schema or program language, 
% is difficult to annotate at scale,
% learning from context-answer pairs $\{(\bx_i, \by_i)\}$ is preferable.

\vspace{-0.2cm}
\subsection{Experimental setup}
\vspace{-0.2cm}
% \subsection{Datasets}

% The dataset, such as \textsc{WikiTableQuestions}~\cite{pasupat2015tables} and \textsc{WebQuestionsSP}~\cite{yih2016webquestionssp}, usually only provides weak supervision through input-output pairs, $\{(x_i, y_i)\}$, without the true program $z_i$. 

{\bf Datasets.}~\textsc{WikiTableQuestions} ~\cite{pasupat2015tables} contains tables extracted from Wikipedia and question-answer pairs about the tables. See Table~\ref{example-qa} as an example. There are 2,108 tables and 18,496 question-answer pairs splitted into train/dev/test set.. We follow the construction in~\cite{pasupat2015tables} for converting a table into a directed graph that can be queried, where rows and cells are converted to graph nodes while column names become labeled directed edges. For the questions, we use string match to identify phrases that appear in the table. We also identify numbers and dates using the CoreNLP annotation released with the dataset. 
%between them (e.g., ``Venue'' connects the first row with ``Finland'').  
%The graph is augmented with  additional  edges \textsc{Next} (from  each row to the next), \textsc{Previous} (from each row to the previous) and \textsc{Index} (from each row to its index number). 
%In addition, 
%we also add \nl{might mislead the user to think that we invented these operations}
%there are normalization edges to the row entity, including \textsc{Column-Number} (from the row to the value of the first number found in the cell), \textsc{Column-Num2} (the second number found in the cell), \textsc{Column-Date} (from the row to the date found in the cell). Depending on the content of the cell, it might have multiple normalized values. For example, if a row has ``3-4'' in its ``Score'' column, it will have a \textsc{Score-Number} edge to the integer 3, a \textsc{Score-Num2} edge to 4, and a \textsc{Score-Date} edge to the date XXXX-03-04.
%
The task is challenging in several aspects. 
First, the tables are taken from Wikipedia and cover a wide range of topics.
Second, %there is no  global schema, so 
at test time, new tables that contain  unseen column names 
%and entities (rows and cells) 
appear. 
Third, the table contents are not normalized as in knowledge-bases like Freebase, so there are noises and ambiguities in the table annotation. 
%For example, some cells contains phrases like ``Beijing, China'' and some cells like ``1996'' can be interpreted as both time or number. 
Last, the semantics are more complex comparing to previous datasets like \textsc{WebQuestionsSP} \cite{yih2016webquestionssp}.
%and require deeper understanding of language compositionality. 
It requires multiple-step reasoning using a large set of functions, including comparisons, superlatives, aggregations, and arithmetic operations~\cite{pasupat2015tables}. See Supplementary Material \ref{dsl} for more details about the functions. 
%A large proportion of the questions requires multiple steps of reasoning to solve. 

\textsc{WikiSQL}~\cite{zhong2017seq2sql} is a recent large scale dataset on learning natural language interfaces for databases. It also uses tables extracted from Wikipedia, but is much larger and is annotated with programs (SQL). There are 24,241 tables and 80,654 question-program pairs splitted into train/dev/test set. Comparing to \textsc{WikiTableQuestions}, the semantics are simpler because the SQLs use fewer operators (column selection, aggregation, and conditions). We perform similar preprocessing as for \textsc{WikiTableQuestions}. Most of the state-of-the-art models relies on question-program pairs for supervised training, while we only use the question-answer pairs for weakly supervised training.  

% \vspace*{-.2cm}
% \subsection{Model architecture}
% \vspace*{-.2cm}
{\bf Model architecture.}~We adopt the Neural Symbolic Machines
framework\cite{liang2017nsm}, which combines (1) a neural
``programmer'', which is a seq2seq model augmented by a key-variable
memory that can translate a natural language utterance to a program as
a sequence of tokens, and (2) a symbolic ``computer'', which is an
Lisp interpreter that implements a domain specific language with
built-in functions and provides code assistance by eliminating
syntactically or semantically invalid choices.

For the Lisp interpreter, we added functions according to
\cite{zhang2017macro,Neelakantan2016LearningAN} for \textsc{WikiTableQuestions} experiments and
used the subset of functions equivalent to column selection,
aggregation, and conditions for \textsc{WikiSQL}. See the Supplementary Material \ref{dsl} for more details about functions used.

We implemented the seq2seq model augmented with key-variable memory
from~\cite{liang2017nsm} in TensorFlow~\cite{abadi2016tensorflow}. Some minor differences are:
(1) we used a bi-directional LSTM for the encoder; (2) we used
two-layer LSTM with skip-connections in both the encoder and decoder.
GloVe~\cite{Pennington2014GloveGV} embeddings are used for the
embedding layer in the encoder and also to create embeddings for
column names by averaging the embeddings of the words in a name. Following~\cite{Neelakantan2016LearningAN,krishnamurthy2017neural},
we also add a binary feature in each step of the encoder, indicating
whether this word is found in the table, and an integer feature for a
column name counting how many of the words in the column name appear
in the question.  For the \textsc{WikiTableQuestions} dataset, we use
the CoreNLP annotation of numbers and dates released with the
dataset. For the \textsc{WikiSQL} dataset, only numbers are used, so
we use a simple parser to identify and parse the numbers in the
questions, and the tables are already preprocessed. The tokens of the
numbers and dates are anonymized as two special tokens <NUM> and
<DATE>.
%We use a two-layer LSTM with skip-connections for both the encoder and decoder. 
The hidden size of the LSTM is $200$. We keep the GloVe embeddings fixed
during training, but project it to $200$ dimensions using a trainable linear
transformation. The same architecture is used for both datasets. 

{\bf Training Details.}~We first apply systematic exploration using a
random policy to discover high-reward programs to warm start the
memory buffer of each example. For
\textsc{WikiTableQuestions}, we generated 50k programs per example using systematic exploration with pruning rules inspired by the grammars from~\cite{zhang2017macro} (see Supplementary \ref{apdx:prune}). 
We apply 0.2 dropout on both encoder and decoder. 
Each batch includes samples from 25 examples.
For experiments on \textsc{WikiSQL}, we generated 1k programs per
example due to computational constraint. 
Because the dataset is much larger, we don't use any regularization. 
Each batch includes samples from 125 examples.
We use distributed sampling for \textsc{WikiTableQuestions}. 
For \textsc{WikiSQL}, due to computational constraints, we truncate each memory buffer to top 5 and then enumerate all 5 programs for training. For both experiments, the samples outside memory buffer are drawn using rejection sampling from 1 on-policy sample per example. 
At inference time, we apply beam search of size 5. 
We evaluate the model periodically on the dev set to select the best model. 
%jb: this was written in 'model details' already.
%We also added two contextual features, one used in encoder to indicate which word appears in
%table, and the other used in the decoder to indicate which column name
%appears in question. They are commonly used to incorporate the
%interaction between the table and the question into the model.
% \jb{Did you
%   decide not to make a big deal of this? Do you want to have an
%   ablation of this component or prefer not to?}  \nl{It might be
%   interesting to compare different amount of systematic explorations
%   10k, 1k, 100 per query.}
We apply a distributed actor-learner architecture for training.
The actors use CPUs to generate new trajectories and push the samples
into a queue.  The learner reads batches of data from the
queue and uses GPU to accelerate training (see Supplementary \ref{apdx:actor-learner}). We use Adam optimizer for
training and the learning rate is $10^{-3}$. 
All the hyperparameters are tuned on the dev set. We train the model for 25k steps on WikiTableQuestions and 15k steps on WikiSQL.

\begin{figure}
%\vspace{-0.2in}
% \vspace{-0.5in}
  %\centering
  %\begin{minipage}{2.5in}
\hspace{-0.1in}
\begin{tabular}{cc}
\includegraphics[width=2.8in]{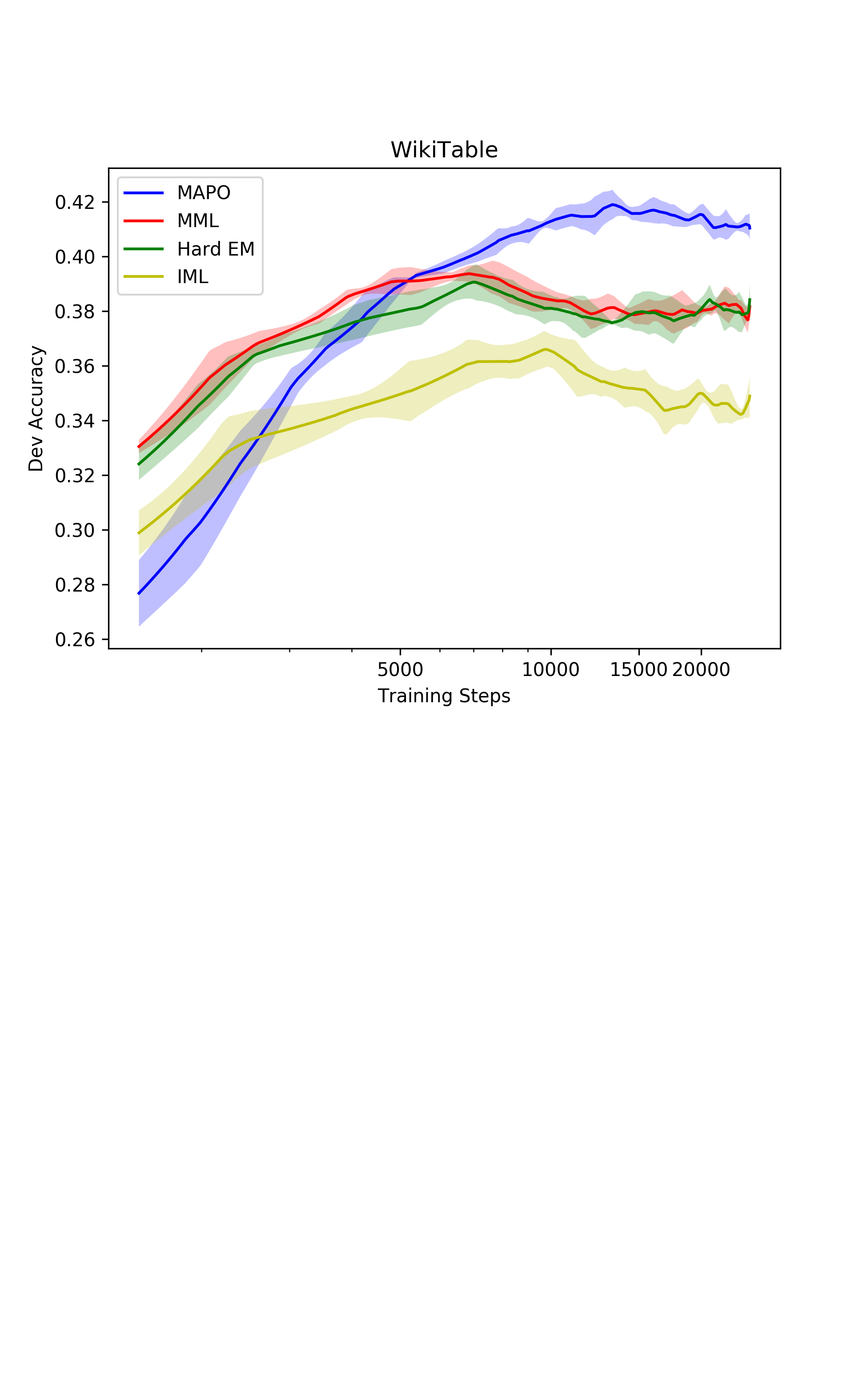}&
\hspace{-0.2in}
\includegraphics[width=2.8in]{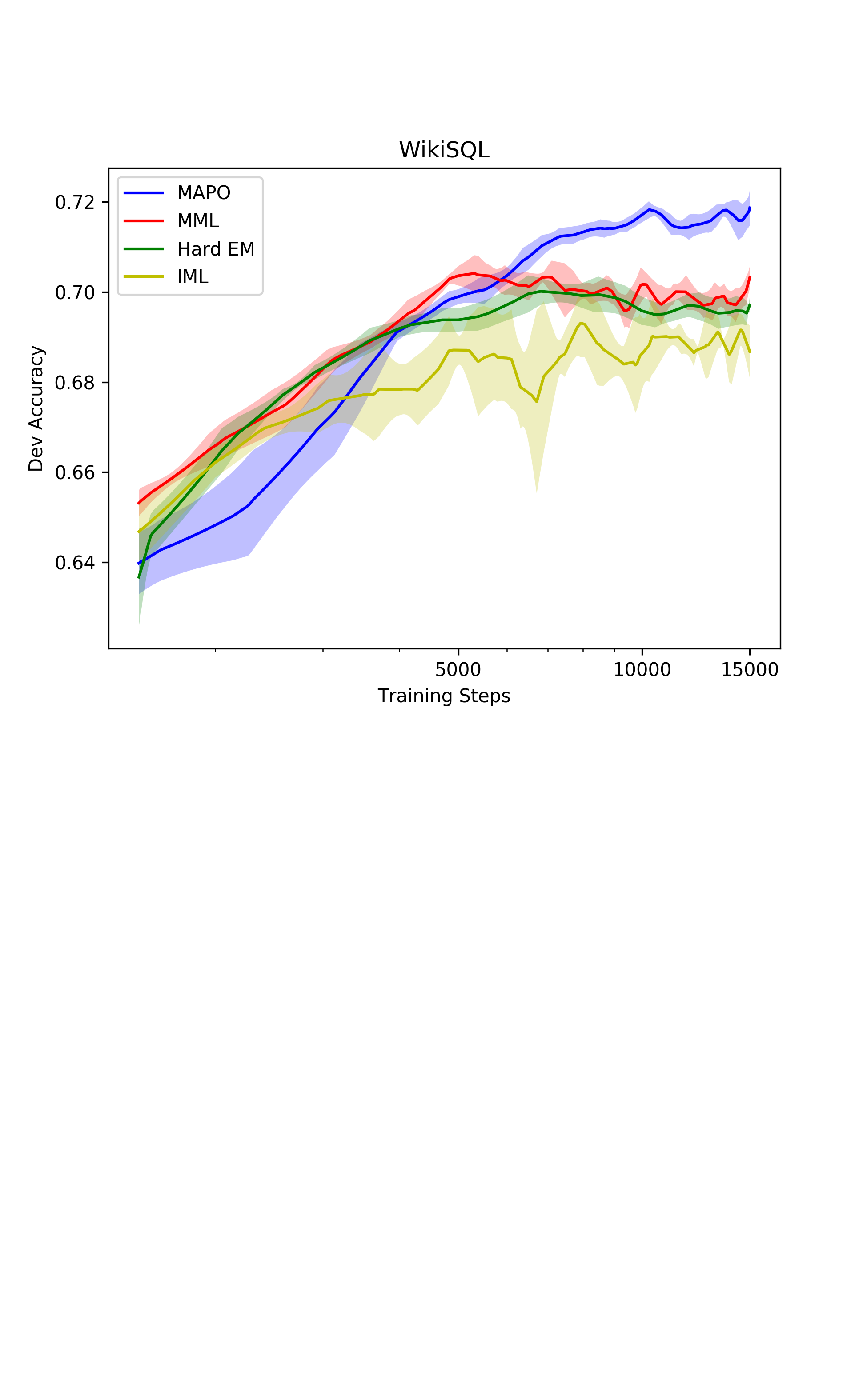}
\end{tabular}
 % \end{minipage}
 \vspace{-2.4in}
  \caption{Comparison of MAPO and 3 baselines' dev set accuracy curves. Results on \textsc{WikiTableQuestions} is on the left and results on \textsc{WikiSQL} is on the right. The plot is average of 5 runs with a bar of one standard deviation. The horizontal coordinate (training steps) is in log scale. 
    % The shaded area represents the standard deviation.
    %The curves are smoothed by a exponentially weighted moving average for clarity.  
    }
  \label{fig:comparison-to-baseline}
\vspace{-0.2in}
\end{figure}

\vspace{-0.2cm}
\subsection{Comparison to baselines %implemented within our framework
}
\vspace{-0.2cm}

We first compare MAPO against the following
baselines using the same neural architecture.\\[.1cm]
%\begin{itemize}[leftmargin=*]
\raisebox{0.1ex}{$\blacktriangleright$} \textbf{REINFORCE:} We use on-policy samples to
estimate the gradient of expected return as in \eqref{eq:pg}, not
utilizing any form of memory buffer.\\[.1cm]
\raisebox{0.1ex}{$\blacktriangleright$} \textbf{MML:} Maximum Marginal
Likelihood maximizes the marginal probability of the memory buffer as
in $\objmml(\theta) = \frac{1}{N} \sum_i \log \sum_{\ba \in
  \mathcal{B}_i} \pit(\ba) = \frac{1}{N} \log \prod_i \sum_{\ba \in
  \mathcal{B}_i} \pit(\ba)$.  Assuming binary rewards and assuming that
the memory buffer contains almost all of the trajectories with a
reward of $1$, MML optimizes the marginal probability of generating a
rewarding program. Note that under these assumptions, expected return
can be expressed as $\objer(\theta) \approx \frac{1}{N}
\sum_i \sum_{\ba \in \buf_i} \pit(\ba)$.
Comparing the two objectives, we can see that
MML maximizes the
product of marginal probabilities, whereas expected return maximizes
the sum. More discussion of these two objectives can be found
in \cite{guu2017language,norouzi2016reward,roux2016tighter}.\\[.1cm]
\raisebox{0.1ex}{$\blacktriangleright$} \textbf{Hard EM:} Expectation-Maximization algorithm is commonly used
to optimize the marginal likelihood in the presence of latent
variables. Hard EM uses the samples with the highest probability to
approximate the gradient to $\objmml$.\\[.1cm]
\raisebox{0.1ex}{$\blacktriangleright$} \textbf{IML:} Iterative
Maximum Likelihood training~\cite{liang2017nsm,abolafia2018neural} uniformly maximizes
the likelihood of all the trajectories with the highest rewards
$\mathcal{O}_{\mathrm{ML}}(\theta) = \sum_{\ba \in \buf} \log \pit(\ba)$.

Because the memory buffer is too large to enumerate, we use samples
from the buffer to approximate the gradient for MML and IML,
and uses samples with highest $\pit(\ba)$ for Hard EM.

We show the result in \tabref{tab:baseline} and the
dev accuracy curves
in \figref{fig:comparison-to-baseline}. Removing systematic
exploration or the memory weight clipping significantly weaken
MAPO because high-reward trajectories are not found or easily
forgotten. REINFORCE barely learns anything because starting
from a random policy, most samples result in a reward of zero. MML and
Hard EM converge faster, but the learned models underperform MAPO,
which suggests that the expected return is a better objective. IML
runs faster because it randomly samples from
the buffer, but the objective is prone to %finding 
spurious programs.
% We can see an interesting 
% trade-off between bias and variance: REINFORCE has no bias but highest variance. It cannot converge within any reasonable time;
% Hard EM and MML have low variances but high bias. They converges faster but to worse solutions;
% {MAPO} introduces some bias and variance. It converges slower but to better solutions. 

%
%
\begin{wraptable}{r}{2.7in}

%\vspace*{-.1in}

%==== Ablation Study ====
  
\setlength{\tabcolsep}{8pt}
\footnotesize 
\vspace{-0.5in}
\centering 
  \begin{tabular}{l@{\hspace{2pt}}cc}
    \toprule
     & \textsc{WikiTable} & \textsc{WikiSQL} \\%(\%) \\ 
    \midrule %\hline
    REINFORCE & $< 10$ & $< 10$ \\
    MML (Soft EM) & $39.7 \pm 0.3$ & $70.7 \pm 0.1$ \\
    Hard EM & $39.3 \pm 0.6$ & $70.2 \pm 0.3$ \\
    IML & $36.8 \pm 0.5$ & $70.1 \pm 0.2$ \\
    %\midrule
    MAPO & ${\bf 42.3}\pm0.3$ & ${\bf 72.2}\pm0.2$ \\
    MAPO w/o SE & $< 10$ & $< 10$ \\
    MAPO w/o MWC  & $< 10$ & $< 10$ \\
    \bottomrule
\end{tabular}
\vspace{-0.05in}
\caption{Ablation study %on both datasets 
for Systematic Exploration (SE) and Memory Weight Clipping (MWC). 
We report mean accuracy \%, %$\pm$ 
and its standard deviation on dev sets based on 5 runs.  }
\label{tab:baseline}

%\vspace*{0.22in}
\vspace*{0.15in}

% ==== WikiTable ===

\setlength{\tabcolsep}{4pt}
\centering
\begin{tabular}{ l ccc}
\toprule
 & E.S. & Dev. & Test \\ 
\midrule %\hline
Pasupat \& Liang (2015)~\cite{pasupat2015tables}  & - & 37.0 & 37.1 \\
Neelakantan \etal~(2017)~\cite{Neelakantan2016LearningAN} & 1 & 34.1 & 34.2  \\
Neelakantan \etal~(2017)~\cite{Neelakantan2016LearningAN} & 15 & 37.5 & 37.7 \\
Haug \etal~(2017)~\cite{haug2018NeuralMR} & 1 & - & 34.8  \\
Haug \etal~(2017)~\cite{haug2018NeuralMR} & 15 & - & 38.7 \\
Zhang \etal~(2017)~\cite{zhang2017macro}%  Macro Grammars 
& - & 40.4 & 43.7 \\ 
%\midrule
MAPO  & 1 & \textbf{42.7} & $43.8$   \\
MAPO (mean of 5 runs) & - & $42.3$ & $43.1$   \\
MAPO (std of 5 runs) & - & $0.3$ & $0.5$   \\
MAPO (ensembled) & 10 & - & {\bf 46.3}  \\
\bottomrule
\end{tabular}
\vspace{-0.05in}
\caption[Results on the dev]{Results on 
  \textsc{WikiTableQuestions}. 
  E.S. is the ensemble size, when applicable.
  %\footnote{We find it more informative to report the mean accuracy and standard deviation based on 5 runs, although it is not available from other models.}
}
\label{tab-wtq}

% \vspace{0.22in}
\vspace{0.1in}

% ==== WikiSQL ====

\setlength{\tabcolsep}{8pt}
\centering 
  \begin{tabular}{lccc}
  	\toprule
   % Method & Dev. & Test \\ 
  %  \midrule % \hline
    \textbf{Fully supervised} &  Dev. & Test \\
    \midrule
    Zhong \etal~(2017)~\cite{zhong2017seq2sql} &  60.8 & 59.4 \\
    Wang \etal~(2017)~\cite{wang2017pointing} &67.1 & 66.8 \\
    Xu \etal~(2017)~\cite{xu2018sqlnet} & 69.8 & 68.0 \\
    Huang \etal~(2018)~\cite{huang2018NaturalLT} & 68.3 & 68.0 \\ 
    Yu \etal~(2018)~\cite{Yu2018TypeSQLKT}& 74.5 & 73.5 \\ 
    Sun \etal~(2018)~\cite{Sun2018SemanticPW} & 75.1 & 74.6 \\ 
    Dong \& Lapata (2018)~\cite{Dong2018CoarsetoFineDF} & \textbf{79.0} & \textbf{78.5} \\
    \midrule
    \midrule
    \textbf{Weakly supervised} & Dev. & Test \\
    \midrule
    MAPO  & \textbf{72.4} & 72.6  \\
    MAPO (mean of 5 runs) & $72.2$ & $72.1$   \\
    MAPO (std of 5 runs) & $0.2$ & $0.3$   \\
    MAPO (ensemble of 10) & - & \textbf{74.9} \\
    \bottomrule
  \end{tabular}
\vspace{-0.05in}
\caption{Results on \textsc{WikiSQL}. 
Unlike other methods, MAPO only uses weak supervision.  
%while all other methods use question-program pairs.
%as full supervision. 
}
\label{tab-wsql}

%\vspace*{-.6in}
\vspace{.6in}

\end{wraptable}
\vspace{-0.2cm}
\subsection{Comparison to state-of-the-art}
%\vspace{-0.2cm}
On \textsc{WikiTableQuestions} (Table~\ref{tab-wtq}), MAPO is the
first RL-based approach that significantly outperforms the previous
state-of-the-art by 2.6\%.  Unlike previous work, MAPO does not
require manual feature engineering or additional human
annotation\footnote{Krishnamurthy \etal~\cite{krishnamurthy2017neural}
achieved 45.9 accuracy when trained on the data collected with dynamic
programming and pruned with more human
annotations~\cite{pasupat2016denotations,mudrakarta2018training}.}. 
On \textsc{WikiSQL} (Table~\ref{tab-wsql}), even though MAPO does not
exploit ground truth programs (weak supervision), it is able to outperform many strong baselines trained using programs (full supervision). The techniques introduced in other
models can be incorporated to further improve the result of MAPO, but
we leave that as future work. 
We also qualitatively analyzed a trained model and see that it can generate fairly complex programs. See the Supplementary Material \ref{example-prog} for some examples of generated programs. 
We select the best model from 5 runs based on validation accuracy and report its test accuracy. We also report the mean accuracy and standard deviation based on 5 runs given the variance caused by the non-linear optimization procedure, although it is not available from other models. 

\subsection{Analysis of Memory Weight Clipping}
\label{sec:expt-clip}

\begin{figure}[t]
% \vspace{-0.2in}
\begin{center}
\begin{tabular}{c@{\hspace*{.5in}}c}
\includegraphics[width=2in]{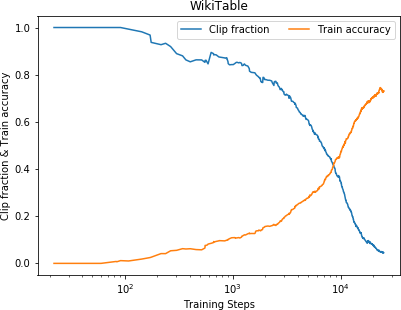}&
\includegraphics[width=2in]{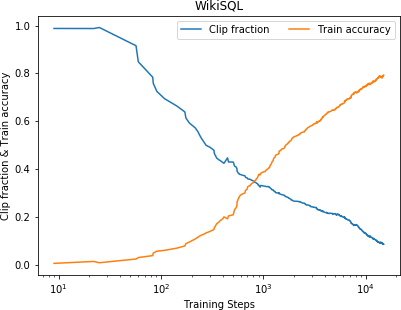}
\end{tabular}
\end{center}
\vspace{-0.1in}
\caption{The clipping fraction and training accuracy w.r.t the training steps (log scale). }
\label{clip}
\label{fig:clip-frac}
\vspace{-0.2in}
\end{figure}

In this subsection, we present an analysis of the bias introduced by memory weight clipping. 
We define the clipping fraction as the percentage of examples where the clipping is active. In other words, it is the percentage of examples with a non-empty memory buffer, for which $\piB < \alpha$.  %the sum of the probabilities of the stored trajectories is smaller than the clipping threshold $\alpha$. 
It is also the fraction of examples whose gradient computation will be biased by the clipping, so the higher the value, the more bias, and the gradient is unbiased when the clip fraction is zero. 
In figure \ref{fig:clip-frac}, one can observe that the clipping fraction approaches zero towards the end of training and is negatively correlated with the training accuracy. 
In the experiments, we found that a fixed clipping threshold works well, but we can also gradually decrease the clipping threshold to completely remove the bias.

\vspace{-0.2cm}
\section{Related work}
\vspace{-0.2cm}

{\bf Program synthesis \& semantic parsing.}~There has been a surge of
recent interest in applying reinforcement learning to program
synthesis~\cite{bunel2018leveraging,abolafia2018neural,zaremba2015reinforcement,nachum2017bridging}
and combinatorial
optimization~\cite{zoph2016neural,bello2016neural}. Different from
these efforts, we focus on the contextualized program synthesis where
generalization to new contexts is important. Semantic
parsing~\cite{zelle96geoquery,zettlemoyer05ccg,liang11dcs} maps
natural language to executable 
symbolic representations.
%that can be executed by an interpreter.
Training semantic parsers through weak supervision is
challenging because the model must interact with a symbolic
interpreter through non-differentiable operations to search over a
large space of
programs \cite{berant2013semantic,liang2017nsm}. Previous
work~\cite{guu2017language,Neelakantan2016LearningAN} reports negative
results when applying simple policy gradient methods like
REINFORCE~\cite{Williams92simplestatistical}, which highlights the
difficulty of exploration and optimization when applying RL techniques.
%to program synthesis or semantic parsing. 
MAPO takes advantage of
discrete and deterministic nature of program synthesis and
significantly improves upon REINFORCE.

{\bf Experience replay.}~An experience replay
buffer~\cite{lin1992self} enables storage and usage of past
experiences to improve the sample efficiency of RL
algorithms. Prioritized experience replay~\cite{schaul2016prioritized}
prioritizes replays based on temporal-difference error for more
efficient optimization. Hindsight experience
replay~\cite{andrychowicz2017hindsight} incorporates goals into replays to deal with sparse rewards. MAPO also uses past experiences to tackle sparse
reward problems, but by storing and reusing high-reward trajectories, similar
to \cite{liang2017nsm, oh2018self}.  
Previous work\cite{liang2017nsm} assigns a fixed weight to the trajectories, which introduces bias into the
policy gradient estimates. More importantly, the policy is often
trained equally on the trajectories that have the same reward, which
is prone to spurious programs. By contrast, MAPO uses the
trajectories in a principled way to obtain an unbiased low variance
gradient estimate.

{\bf Variance reduction.}~Policy optimization via gradient descent is
challenging because of: \mbox{(1)}~large {\em variance} in gradient
estimates;
\mbox{(2)}~small gradients in the initial phase of training.  Prior
variance reduction
approaches~\cite{wu2018variance,Williams92simplestatistical,liu2017sample,grathwohl2017backpropagation} mainly
relied on control variate techniques by introducing a
critic model~\cite{konda2000actor,mnih2016asynchronous,ppo2017}.  MAPO
takes a different approach to reformulate the gradient as a combination
of expectations inside and outside a memory buffer. Standard solutions
to the small gradient problem involves supervised
pretraining~\cite{silver2016mastering,hester2017demonstrations,ranzato2015sequence}
or using supervised data to generate rewarding
samples~\cite{norouzi2016reward,ding2017cold}, which cannot be applied
when supervised data are not available. MAPO reduces the variance by sampling from a smaller stochastic space or through stratified sampling, and
accelerates and stabilizes training by clipping the  weight of the memory buffer. 

{\bf Exploration.}~Recently there has been a lot of work on improving exploration
\cite{pathak2017curiosity,tang2017count,houthooft2016vime} by introducing additional reward based on information gain or pseudo count.
For program
synthesis~\cite{balog2017deepcoder,Neelakantan2016LearningAN,bunel2018leveraging},
the search spaces are enumerable and deterministic.  Therefore, we
propose to conduct systematic exploration, which ensures that only novel
trajectories are generated.

\vspace{-0.2cm}
\section{Conclusion}
\vspace{-0.2cm} We present memory augmented policy optimization
(\MAPO) that incorporates a memory buffer of promising 
trajectories to reduce the variance of policy gradients. We propose 3
techniques to enable an efficient algorithm for \MAPO: (1) memory weight clipping to accelerate and stabilize training; (2) systematic exploration to efficiently discover high-reward trajectories;
(3)
distributed sampling from inside and outside memory buffer to scale up training.
%We show that \MAPO outperforms several baselines. 
\MAPO is evaluated on real world program synthesis from natural language / semantic parsing tasks. On \textsc{WikiTableQuestions}, \MAPO is the first RL approach that significantly outperforms previous state-of-the-art; on \textsc{WikiSQL}, \MAPO trained with only weak supervision outperforms several strong baselines trained with full supervision. 

\vspace{-0.2cm}
\subsubsection*{Acknowledgments}
\vspace{-0.2cm}
We would like to thank Dan Abolafia, Ankur Taly,
Thanapon Noraset, Arvind Neelakantan, Wenyun Zuo, Chenchen Pan and Mia Liang for helpful discussions.  Jonathan Berant was partially
supported by The Israel Science Foundation grant 942/16.

\bibliographystyle{plain}
\bibliography{bib.bib}

\begin{thebibliography}{10}

\bibitem{abadi2016tensorflow}
Mart{\'{\i}}n Abadi, Ashish Agarwal, Paul Barham, Eugene Brevdo, Zhifeng Chen,
  Craig Citro, Gregory~S. Corrado, Andy Davis, Jeffrey Dean, Matthieu Devin,
  Sanjay Ghemawat, Ian~J. Goodfellow, Andrew Harp, Geoffrey Irving, Michael
  Isard, Yangqing Jia, Rafal J{\'{o}}zefowicz, Lukasz Kaiser, Manjunath Kudlur,
  Josh Levenberg, Dan Man{\'{e}}, Rajat Monga, Sherry Moore, Derek~Gordon
  Murray, Chris Olah, Mike Schuster, Jonathon Shlens, Benoit Steiner, Ilya
  Sutskever, Kunal Talwar, Paul~A. Tucker, Vincent Vanhoucke, Vijay Vasudevan,
  Fernanda~B. Vi{\'{e}}gas, Oriol Vinyals, Pete Warden, Martin Wattenberg,
  Martin Wicke, Yuan Yu, and Xiaoqiang Zheng.
\newblock Tensorflow: Large-scale machine learning on heterogeneous distributed
  systems.
\newblock {\em ArXiv:1603.04467}, 2016.

\bibitem{abolafia2018neural}
Daniel~A Abolafia, Mohammad Norouzi, and Quoc~V Le.
\newblock Neural program synthesis with priority queue training.
\newblock {\em arXiv preprint arXiv:1801.03526}, 2018.

\bibitem{pqt2018}
Daniel~A. Abolafia, Mohammad Norouzi, Jonathan Shen, Rui Zhao, and Quoc~V. Le.
\newblock Neural program synthesis with priority queue training.
\newblock {\em arXiv:1801.03526}, 2018.

\bibitem{andrychowicz2017hindsight}
Marcin Andrychowicz, Filip Wolski, Alex Ray, Jonas Schneider, Rachel Fong,
  Peter Welinder, Bob McGrew, Josh Tobin, OpenAI~Pieter Abbeel, and Wojciech
  Zaremba.
\newblock Hindsight experience replay.
\newblock {\em NIPS}, 2017.

\bibitem{balog2017deepcoder}
M.~Balog, A.~L. Gaunt, M.~Brockschmidt, S.~Nowozin, and D.~Tarlow.
\newblock Deepcoder: Learning to write programs.
\newblock {\em ICLR}, 2017.

\bibitem{arcade2013}
Marc~G Bellemare, Yavar Naddaf, Joel Veness, and Michael Bowling.
\newblock The arcade learning environment: An evaluation platform for general
  agents.
\newblock {\em JMLR}, 2013.

\bibitem{bello2016neural}
Irwan Bello, Hieu Pham, Quoc~V Le, Mohammad Norouzi, and Samy Bengio.
\newblock Neural combinatorial optimization with reinforcement learning.
\newblock {\em arXiv:1611.09940}, 2016.

\bibitem{berant2013semantic}
Jonathan Berant, Andrew Chou, Roy Frostig, and Percy Liang.
\newblock Semantic parsing on freebase from question-answer pairs.
\newblock {\em EMNLP}, 2(5):6, 2013.

\bibitem{gym2016}
Greg Brockman, Vicki Cheung, Ludwig Pettersson, Jonas Schneider, John Schulman,
  Jie Tang, and Wojciech Zaremba.
\newblock Openai gym.
\newblock {\em arXiv:1606.01540}, 2016.

\bibitem{bunel2018leveraging}
Rudy Bunel, Matthew Hausknecht, Jacob Devlin, Rishabh Singh, and Pushmeet
  Kohli.
\newblock Leveraging grammar and reinforcement learning for neural program
  synthesis.
\newblock In {\em International Conference on Learning Representations}, 2018.

\bibitem{dasdialog2017}
Abhishek Das, Satwik Kottur, Jos{\'e}~MF Moura, Stefan Lee, and Dhruv Batra.
\newblock Learning cooperative visual dialog agents with deep reinforcement
  learning.
\newblock {\em arXiv:1703.06585}, 2017.

\bibitem{degris2012}
Thomas Degris, Martha White, and Richard~S Sutton.
\newblock Off-policy actor-critic.
\newblock {\em ICML}, 2012.

\bibitem{ding2017cold}
Nan Ding and Radu Soricut.
\newblock Cold-start reinforcement learning with softmax policy gradient.
\newblock In {\em Advances in Neural Information Processing Systems}, pages
  2817--2826, 2017.

\bibitem{Dong2018CoarsetoFineDF}
Li~Dong and Mirella Lapata.
\newblock Coarse-to-fine decoding for neural semantic parsing.
\newblock {\em CoRR}, abs/1805.04793, 2018.

\bibitem{espeholt2018impala}
Lasse Espeholt, Hubert Soyer, Remi Munos, Karen Simonyan, Volodymir Mnih, Tom
  Ward, Yotam Doron, Vlad Firoiu, Tim Harley, Iain Dunning, et~al.
\newblock Impala: Scalable distributed deep-rl with importance weighted
  actor-learner architectures.
\newblock {\em arXiv:1802.01561}, 2018.

\bibitem{grathwohl2017backpropagation}
Will Grathwohl, Dami Choi, Yuhuai Wu, Geoff Roeder, and David Duvenaud.
\newblock Backpropagation through the void: Optimizing control variates for
  black-box gradient estimation.
\newblock {\em arXiv preprint arXiv:1711.00123}, 2017.

\bibitem{guu2017language}
Kelvin Guu, Panupong Pasupat, Evan Liu, and Percy Liang.
\newblock From language to programs: Bridging reinforcement learning and
  maximum marginal likelihood.
\newblock {\em ACL}, 2017.

\bibitem{haug2018NeuralMR}
Till Haug, Octavian-Eugen Ganea, and Paulina Grnarova.
\newblock Neural multi-step reasoning for question answering on semi-structured
  tables.
\newblock In {\em ECIR}, 2018.

\bibitem{hester2017demonstrations}
Todd Hester, Matej Vecerik, Olivier Pietquin, Marc Lanctot, Tom Schaul, Bilal
  Piot, Andrew Sendonaris, Gabriel Dulac-Arnold, Ian Osband, John Agapiou, Joel
  Z.~Leibo, and Audrunas Gruslys.
\newblock Deep q-learning from demonstrations.
\newblock {\em AAAI}, 2018.

\bibitem{hochreiter1997long}
Sepp Hochreiter and J{\"u}rgen Schmidhuber.
\newblock Long short-term memory.
\newblock {\em Neural Comput.}, 1997.

\bibitem{houthooft2016vime}
Rein Houthooft, Xi~Chen, Yan Duan, John Schulman, Filip De~Turck, and Pieter
  Abbeel.
\newblock Vime: Variational information maximizing exploration.
\newblock In {\em Advances in Neural Information Processing Systems}, pages
  1109--1117, 2016.

\bibitem{huang2018NaturalLT}
Po-Sen Huang, Chenglong Wang, Rishabh Singh, Wen tau Yih, and Xiaodong He.
\newblock Natural language to structured query generation via meta-learning.
\newblock {\em CoRR}, abs/1803.02400, 2018.

\bibitem{konda2000actor}
Vijay~R Konda and John~N Tsitsiklis.
\newblock Actor-critic algorithms.
\newblock In {\em Advances in neural information processing systems}, pages
  1008--1014, 2000.

\bibitem{krishnamurthy2017neural}
Jayant Krishnamurthy, Pradeep Dasigi, and Matt Gardner.
\newblock Neural semantic parsing with type constraints for semi-structured
  tables.
\newblock {\em EMNLP}, 2017.

\bibitem{li2016deep}
Jiwei Li, Will Monroe, Alan Ritter, Michel Galley, Jianfeng Gao, and Dan
  Jurafsky.
\newblock Deep reinforcement learning for dialogue generation.
\newblock {\em arXiv:1606.01541}, 2016.

\bibitem{liang2017nsm}
Chen Liang, Jonathan Berant, Quoc Le, Kenneth~D. Forbus, and Ni~Lao.
\newblock Neural symbolic machines: Learning semantic parsers on freebase with
  weak supervision.
\newblock {\em ACL}, 2017.

\bibitem{liang11dcs}
P.~Liang, M.~I. Jordan, and D.~Klein.
\newblock Learning dependency-based compositional semantics.
\newblock {\em ACL}, 2011.

\bibitem{lin1992self}
Long-Ji Lin.
\newblock Self-improving reactive agents based on reinforcement learning,
  planning and teaching.
\newblock {\em Machine learning}, 8(3-4):293--321, 1992.

\bibitem{liu2017sample}
Hao Liu, Yihao Feng, Yi~Mao, Dengyong Zhou, Jian Peng, and Qiang Liu.
\newblock Sample-efficient policy optimization with stein control variate.
\newblock {\em arXiv preprint arXiv:1710.11198}, 2017.

\bibitem{rs2018}
Horia Mania, Aurelia Guy, and Benjamin Recht.
\newblock Simple random search provides a competitive approach to reinforcement
  learning.
\newblock {\em arXiv preprint arXiv:1803.07055}, 2018.

\bibitem{mnih2016asynchronous}
Volodymyr Mnih, Adria~Puigdomenech Badia, Mehdi Mirza, Alex Graves, Timothy
  Lillicrap, Tim Harley, David Silver, and Koray Kavukcuoglu.
\newblock Asynchronous methods for deep reinforcement learning.
\newblock In {\em International Conference on Machine Learning}, pages
  1928--1937, 2016.

\bibitem{mudrakarta2018training}
Pramod~Kaushik Mudrakarta, Ankur Taly, Mukund Sundararajan, and Kedar
  Dhamdhere.
\newblock It was the training data pruning too!
\newblock {\em arXiv:1803.04579}, 2018.

\bibitem{nachum2017bridging}
Ofir Nachum, Mohammad Norouzi, Kelvin Xu, and Dale Schuurmans.
\newblock Bridging the gap between value and policy based reinforcement
  learning.
\newblock In {\em Advances in Neural Information Processing Systems}, pages
  2775--2785, 2017.

\bibitem{Neelakantan2016LearningAN}
Arvind Neelakantan, Quoc~V. Le, Mart{\'i}n Abadi, Andrew~D McCallum, and Dario
  Amodei.
\newblock Learning a natural language interface with neural programmer.
\newblock {\em arXiv:1611.08945}, 2016.

\bibitem{gotta2018gotta}
Alex Nichol, Vicki Pfau, Christopher Hesse, Oleg Klimov, and John Schulman.
\newblock Gotta learn fast: A new benchmark for generalization in rl.
\newblock {\em arXiv:1804.03720}, 2018.

\bibitem{norouzi2016reward}
Mohammad Norouzi, Samy Bengio, Navdeep Jaitly, Mike Schuster, Yonghui Wu, Dale
  Schuurmans, et~al.
\newblock Reward augmented maximum likelihood for neural structured prediction.
\newblock In {\em Advances In Neural Information Processing Systems}, pages
  1723--1731, 2016.

\bibitem{nowozin2011structured}
Sebastian Nowozin, Christoph~H Lampert, et~al.
\newblock Structured learning and prediction in computer vision.
\newblock {\em Foundations and Trends{\textregistered} in Computer Graphics and
  Vision}, 6(3--4):185--365, 2011.

\bibitem{oh2018self}
Junhyuk Oh, Yijie Guo, Satinder Singh, and Honglak Lee.
\newblock Self-imitation learning.
\newblock {\em ICML}, 2018.

\bibitem{pasupat2015tables}
Panupong Pasupat and Percy Liang.
\newblock Compositional semantic parsing on semi-structured tables.
\newblock {\em ACL}, 2015.

\bibitem{pasupat2016inferring}
Panupong Pasupat and Percy Liang.
\newblock Inferring logical forms from denotations.
\newblock In {\em Proceedings of the 54th Annual Meeting of the Association for
  Computational Linguistics (Volume 1: Long Papers)}, volume~1, pages 23--32,
  2016.

\bibitem{pasupat2016denotations}
Panupong Pasupat and Percy Liang.
\newblock Inferring logical forms from denotations.
\newblock {\em ACL}, 2016.

\bibitem{pathak2017curiosity}
Deepak Pathak, Pulkit Agrawal, Alexei~A. Efros, and Trevor Darrell.
\newblock Curiosity-driven exploration by self-supervised prediction.
\newblock In {\em ICML}, 2017.

\bibitem{Pennington2014GloveGV}
Jeffrey Pennington, Richard Socher, and Christopher~D. Manning.
\newblock Glove: Global vectors for word representation.
\newblock {\em EMNLP}, 2014.

\bibitem{peters2006}
Jan Peters and Stefan Schaal.
\newblock Policy gradient methods for robotics.
\newblock {\em IROS}, 2006.

\bibitem{rajeswaran2017towards}
Aravind Rajeswaran, Kendall Lowrey, Emanuel~V Todorov, and Sham~M Kakade.
\newblock Towards generalization and simplicity in continuous control.
\newblock {\em NIPS}, 2017.

\bibitem{ranzato2015sequence}
Marc'Aurelio Ranzato, Sumit Chopra, Michael Auli, and Wojciech Zaremba.
\newblock Sequence level training with recurrent neural networks.
\newblock {\em ICLR}, 2016.

\bibitem{ross2011reduction}
St{\'e}phane Ross, Geoffrey Gordon, and Drew Bagnell.
\newblock A reduction of imitation learning and structured prediction to
  no-regret online learning.
\newblock In {\em Proceedings of the fourteenth international conference on
  artificial intelligence and statistics}, pages 627--635, 2011.

\bibitem{roux2016tighter}
Nicolas~Le Roux.
\newblock Tighter bounds lead to improved classifiers.
\newblock {\em ICLR}, 2017.

\bibitem{schaul2016prioritized}
Tom Schaul, John Quan, Ioannis Antonoglou, and David Silver.
\newblock Prioritized experience replay.
\newblock {\em ICLR}, 2016.

\bibitem{trpo2015}
John Schulman, Sergey Levine, Pieter Abbeel, Michael Jordan, and Philipp
  Moritz.
\newblock Trust region policy optimization.
\newblock {\em ICML}, 2015.

\bibitem{ppo2017}
John Schulman, Filip Wolski, Prafulla Dhariwal, Alec Radford, and Oleg Klimov.
\newblock Proximal policy optimization algorithms.
\newblock {\em arXiv:1707.06347}, 2017.

\bibitem{silver2016mastering}
David Silver, Aja Huang, Chris~J Maddison, Arthur Guez, Laurent Sifre, George
  Van Den~Driessche, Julian Schrittwieser, Ioannis Antonoglou, Veda
  Panneershelvam, Marc Lanctot, et~al.
\newblock Mastering the game of go with deep neural networks and tree search.
\newblock {\em Nature}, 529(7587):484--489, 2016.

\bibitem{silver2017mastering}
David Silver, Julian Schrittwieser, Karen Simonyan, Ioannis Antonoglou, Aja
  Huang, Arthur Guez, Thomas Hubert, Lucas Baker, Matthew Lai, Adrian Bolton,
  et~al.
\newblock Mastering the game of go without human knowledge.
\newblock {\em Nature}, 2017.

\bibitem{Sun2018SemanticPW}
Yibo Sun, Duyu Tang, Nan Duan, Jianshu Ji, Guihong Cao, Xiaocheng Feng, Bing
  Qin, Ting Liu, and Ming Zhou.
\newblock Semantic parsing with syntax-and table-aware sql generation.
\newblock {\em arXiv:1804.08338}, 2018.

\bibitem{tang2017count}
Haoran Tang, Rein Houthooft, Davis Foote, Adam Stooke, OpenAI Xi~Chen, Yan
  Duan, John Schulman, Filip DeTurck, and Pieter Abbeel.
\newblock \#exploration: A study of count-based exploration for deep
  reinforcement learning.
\newblock In I.~Guyon, U.~V. Luxburg, S.~Bengio, H.~Wallach, R.~Fergus,
  S.~Vishwanathan, and R.~Garnett, editors, {\em Advances in Neural Information
  Processing Systems 30}, pages 2753--2762. Curran Associates, Inc., 2017.

\bibitem{wang2017pointing}
Chenglong Wang, Marc Brockschmidt, and Rishabh Singh.
\newblock Pointing out {SQL} queries from text.
\newblock {\em ICLR}, 2018.

\bibitem{wang2016sample}
Ziyu Wang, Victor Bapst, Nicolas Heess, Volodymyr Mnih, Remi Munos, Koray
  Kavukcuoglu, and Nando de~Freitas.
\newblock Sample efficient actor-critic with experience replay.
\newblock {\em ICLR}, 2017.

\bibitem{Williams92simplestatistical}
Ronald~J. Williams.
\newblock Simple statistical gradient-following algorithms for connectionist
  reinforcement learning.
\newblock {\em Machine Learning}, pages 229--256, 1992.

\bibitem{wu2018variance}
Cathy Wu, Aravind Rajeswaran, Yan Duan, Vikash Kumar, Alexandre~M Bayen, Sham
  Kakade, Igor Mordatch, and Pieter Abbeel.
\newblock Variance reduction for policy gradient with action-dependent
  factorized baselines.
\newblock {\em ICLR}, 2018.

\bibitem{wu2016gnmt}
Yonghui Wu, Mike Schuster, Zhifeng Chen, Quoc~V. Le, Mohammad Norouzi, Wolfgang
  Macherey, Maxim Krikun, Yuan Cao, Qin Gao, Klaus Macherey, Jeff Klingner,
  Apurva Shah, Melvin Johnson, Xiaobing Liu, Lukasz Kaiser, Stephan Gouws,
  Yoshikiyo Kato, Taku Kudo, Hideto Kazawa, Keith Stevens, George Kurian,
  Nishant Patil, Wei Wang, Cliff Young, Jason Smith, Jason Riesa, Alex Rudnick,
  Oriol Vinyals, Greg Corrado, Macduff Hughes, and Jeffrey Dean.
\newblock Google's neural machine translation system: Bridging the gap between
  human and machine translation.
\newblock {\em arXiv:1609.08144}, 2016.

\bibitem{xu2018sqlnet}
Xiaojun Xu, Chang Liu, and Dawn Song.
\newblock {SQLN}et: Generating structured queries from natural language without
  reinforcement learning.
\newblock {\em ICLR}, 2018.

\bibitem{yih2016webquestionssp}
Wen-tau Yih, Matthew Richardson, Chris Meek, Ming-Wei Chang, and Jina Suh.
\newblock The value of semantic parse labeling for knowledge base question
  answering.
\newblock {\em ACL}, 2016.

\bibitem{Yu2018TypeSQLKT}
Tao Yu, Zifan Li, Zilin Zhang, Rui Zhang, and Dragomir Radev.
\newblock Typesql: Knowledge-based type-aware neural text-to-sql generation.
\newblock {\em arXiv:1804.09769}, 2018.

\bibitem{zaremba2015reinforcement}
Wojciech Zaremba and Ilya Sutskever.
\newblock Reinforcement learning neural turing machines.
\newblock {\em arXiv:1505.00521}, 2015.

\bibitem{zelle96geoquery}
M.~Zelle and R.~J. Mooney.
\newblock Learning to parse database queries using inductive logic programming.
\newblock {\em Association for the Advancement of Artificial Intelligence
  (AAAI)}, pages 1050--1055, 1996.

\bibitem{zettlemoyer05ccg}
L.~S. Zettlemoyer and M.~Collins.
\newblock Learning to map sentences to logical form: Structured classification
  with probabilistic categorial grammars.
\newblock {\em Uncertainty in Artificial Intelligence (UAI)}, pages 658--666,
  2005.

\bibitem{zhang2017macro}
Yuchen Zhang, Panupong Pasupat, and Percy Liang.
\newblock Macro grammars and holistic triggering for efficient semantic
  parsing.
\newblock {\em ACL}, 2017.

\bibitem{zhong2017seq2sql}
Victor Zhong, Caiming Xiong, and Richard Socher.
\newblock Seq2sql: Generating structured queries from natural language using
  reinforcement learning.
\newblock {\em arXiv:1709.00103}, 2017.

\bibitem{zoph2016}
Barret Zoph and Quoc~V Le.
\newblock Neural architecture search with reinforcement learning.
\newblock {\em ICLR}, 2016.

\bibitem{zoph2016neural}
Barret Zoph and Quoc~V Le.
\newblock Neural architecture search with reinforcement learning.
\newblock {\em arXiv:1611.01578}, 2016.

\bibitem{zoph2017}
Barret Zoph, Vijay Vasudevan, Jonathon Shlens, and Quoc~V Le.
\newblock Learning transferable architectures for scalable image recognition.
\newblock {\em arXiv:1707.07012}, 2017.

\end{thebibliography}
\appendix
\newpage
%\section{Supplemental Material}
%\label{sec:supplemental}

\section{Domain Specific Language}
\label{sec:supplemental}
\label{dsl}

We adopt a Lisp-like domain specific language with certain built-in functions. 
A program $C$ is a list of expressions $(c_1 ... c_N)$, where
each expression is either a special token ``\textit{EOS}" indicating the end of the program,
or a list of tokens enclosed by parentheses ``$( F A_1 ... A_K )$". $F$ is a function,
which takes as input $K$ arguments of specific types. Table~\ref{tab-functions} defines the arguments, return value and semantics of each function. 
In the table domain, there are rows and columns. The value of the table cells can be number, date time or string, so we also categorize the columns into number columns, date time columns and string columns depending on the type of the cell values in the column. 
\begin{table}[!htbp]
%\setlength{\tabcolsep}{2pt}
%\small
\centering 
\begin{tabular}{llll}
\toprule
\bf{Function} & \bf{Arguments} & \bf{Returns} & \bf{Description}  \\
\midrule 
(\textbf{hop} v p) & \textbf{v}: a list of rows.  & a list of cells. & Select the given column of the  \\ 
& \textbf{p}: a column. & & given rows. \\
\midrule 
(\textbf{argmax} v p) & \textbf{v}: a list of rows.  & a list of rows. & From the given rows, select the  \\ 
(\textbf{argmin} v p) & \textbf{p}: a number or date & & ones with the largest / smallest   \\
&   column. & & value in the given column.  \\
% \midrule 
% (\textbf{filter$_=$} v q p)  & \textbf{v}: a list of rows.  & a list of rows. & From the given rows, select the  \\ 
% (\textbf{filter$_{\neq}$} v q p) & \textbf{q}: a number or date time. & & ones whose given column is equal \\
% & \textbf{p}: a number or date time & &  / not equal to the given value. \\
% &  column. & & \\
\midrule 
(\textbf{filter$_>$} v q p)  & \textbf{v}: a list of rows.  & a list of rows. & From the given rows, select the  ones \\ 
(\textbf{filter$_{\geq}$} v q p) & \textbf{q}: a number or date. & & whose given column has  certain  \\
(\textbf{filter$_<$} v q p) & \textbf{p}: a number or date & & order relation with the given value. \\
%greater / greater equal / less / less equal \\
(\textbf{filter$_{\leq}$} v q p) &   column. & & 
\\ %than / equal/ not equal the given value.  \\
(\textbf{filter$_=$} v q p)  & &&\\
(\textbf{filter$_{\neq}$} v q p) &&&\\
\midrule 
(\textbf{filter$_{in}$} v q p) & \textbf{v}: a list of rows.  & a list of rows. & From the given rows, select the \\ 
(\textbf{filter$_{!in}$} v q p)  & \textbf{q}: a string. & & ones whose given column contain \\
& \textbf{p}: a string column. & & / do not contain the given string.  \\
\midrule 
(\textbf{first} v) & \textbf{v}: a list of rows.  & a row. & From the given rows, select the one \\ 
(\textbf{last} v) & & & with the smallest / largest index.  \\
\midrule 
(\textbf{previous} v) & \textbf{v}: a row.  & a row. & Select the row that is above \\ 
(\textbf{next} v) & & &  / below the given row.   \\
\midrule 
(\textbf{count} v) & \textbf{v}: a list of rows.  & a number. & Count the number of given rows. \\ 
\midrule 
(\textbf{max} v p) & \textbf{v}: a list of rows.  & a number. & Compute the maximum / minimum \\ 
(\textbf{min} v p) & \textbf{p}: a number column. & & / average / sum of the given column \\
(\textbf{average} v p) & & &   in the given rows. \\
(\textbf{sum} v p) &  & &  \\
\midrule 
(\textbf{mode} v p) & \textbf{v}: a list of rows.  & a cell. & Get the most common value of the  \\
& \textbf{p}: a column. & & given column in the given rows. \\
\midrule 
(\textbf{same\_as} v p) & \textbf{v}: a row. & a list of rows. & Get the rows whose given column is \\
& \textbf{p}: a column. & & the same as the given row.  \\
\midrule 
(\textbf{diff} v0 v1 p) & \textbf{v0}: a row.  & a number. & Compute the difference in the given \\
 & \textbf{v1}: a row. & & column of the given two rows. \\
  & \textbf{p}: a number column. & &  \\
\bottomrule
\end{tabular}
\label{tab-functions}
\setlength{\abovecaptionskip}{10pt}
\caption{Functions used in the experiments.}
\vspace{-0.2in}
\end{table}

In the \textsc{WikiTableQuestions} experiments, we used all the functions in the table. In the \textsc{WikiSQL} experiments, because the semantics of the questions are simpler, we used a subset of the functions (hop, filter$_=$, filter$_{in}$, filter$_>$, filter$_<$, count, maximum, minimum, average and sum). We created the functions according to \cite{zhang2017macro,Neelakantan2016LearningAN}.\footnote{The only function we have added to capture some complex semantics is the same\_as function, but it only appears in $~1.2\%$ of the generated programs (among which $~0.6\%$ are correct and the other $~0.6\%$ are incorrect), so even if we remove it, the significance of the difference in Table \ref{tab-wtq} will not change. }

\section{Examples of Generated Programs}
\label{example-prog}

The following table shows examples of several types of programs generated by a trained model.
%one example solution for each type of task.
%Only the tuple with the highest probability is shown for each sentence.

\begin{table}[!htbp]
\setlength{\tabcolsep}{2pt}
%\scriptsize
\footnotesize %\small
\centering 
\begin{tabular}{lll}
\toprule
%\bf{Type} 
& \bf{Statement} & \bf{Comment}  \\
\midrule 
\midrule
& \textbf{Superlative} \\
\hline
& \multicolumn{2}{l}{\bf{
nt-13901: the most points were scored by which player?}} \\
%\midrule 
& (argmax all\_rows r.points-num) &
Sort all rows by column `points' and take the first row.  \\
&(hop v0 r.player-str) &
Output the value of column `player' for the rows in v0. \\
\midrule
\midrule
&\textbf{Difference}\\
\hline
& \multicolumn{2}{l}{\bf{
nt-457: how many more passengers flew to los angeles than to saskatoon?}}\\
%\midrule 
&(filter$_{in}$ all\_rows ['saskatoon'] r.city-str)  &
 Find the row with `saskatoon' matched in column `city'.\\
&(filter$_{in}$ all\_rows [`los angeles'] r.city-str)  &
 Find the row with `los angeles' matched in column `city'.\\
&(diff v1 v0 r.passengers-num) &
Calculate the difference of the values \\
&& in the column `passenger' of v0 and v1. \\
\midrule
\midrule 
&\textbf{Before / After}\\ 
\hline
& \multicolumn{2}{l}{\bf{
nt-10832: which nation is before peru?}} \\
& (filter$_{in}$ all\_rows [`peru'] r.nation-str) &
Find the row with `peru' matched in `nation' column. \\
&(previous v0)  &  Find the row before v0. \\
&(hop v1 r.nation-str) & Output the value of column `nation' of v1. \\
\midrule
\midrule 
&\textbf{Compare \& Count}\\ 
\hline
& \multicolumn{2}{l}{\bf{
nt-647: in how many games did sri lanka score at least 2 goals? }} \\
&(filter$_{\geq}$ all\_rows [2] r.score-num)  & Select the rows whose value in the `score' column >= 2. \\
&(count v0)  & Count the number of rows in v0. \\
\midrule
\midrule 
&\textbf{Exclusion}\\
\hline
&\multicolumn{2}{l}{\bf{
nt-1133: other than william stuart price, which other businessman was born in tulsa?}} \\
&(filter$_{in}$ all\_rows [`tulsa'] r.hometown-str) & Find rows with `tulsa' matched in column `hometown'. \\
&(filter$_{!in}$ v0 [`william stuart price'] r.name-str) & Drop rows with `william stuart price'  matched in the\\
& & value of column `name'.\\
&(hop v1 r.name-str) & Output the value of column `name' of v1. \\
% \midrule 
% \midrule 
% &\textbf{Equal}\\
% \hline
% &\multicolumn{2}{l}{\bf{
% nt-601: which other ship was launched in the same year as the wave victor? }}\\
% &(filter$_{in}$ all\_rows [`wave victor'] r.name-str) &
%  Find rows with `wave victor' matched in column `name'.\\
% &(same\_as v0 r.launched-num2)&
%  Find rows with the same value of column `launched' as v0. \\
% &(hop v1 r.name-str) &  Output the value of column `name' of v1.\\
\bottomrule
\end{tabular}
\setlength{\abovecaptionskip}{10pt}
\caption{Example programs generated by a trained model.}
\end{table}

\
\section{Analysis of Sampling from Inside and Outside Memory Buffer}

In the following we give theoretical analysis of the distributed sampling approaches.
For the purpose of the analysis we assume binary rewards, and exhaustive exploration, that the buffer $\buf^+ ~\equiv~\buf$ contains all the high reward samples, and  $\buf^-~\equiv~\A-\Ba$ contains all the non-rewarded samples.
It provides 
%analysis of how the variance of gradient estimations varies according to different sampling strategies, 
a general guidance of how examples should be allocated on the experiences and whether to use baseline or not so that the variance of gradient estimations can be minimized.
In our work, we take the simpler approach to not use baseline and leave the empirical investigation to future work.

%Note that under the assumption of binary rewards and sufficient exploration), we have $\pit(\mathcal{B}) \approx \expected_{\ba \sim \pit(\ba)} R(\ba)$. 

\subsection{Variance: baseline vs no baseline}
\label{apdx:baseline}
Here we compare baseline strategies based on their variances of gradient estimations.
%s when applying stratified sampling estimator to policy gradient. 
We thank Wenyun Zuo's suggestion in approximating the variances.

Assume $\sigma_+^2 = Var_{\ba \sim \pit^+(\ba)}[\nabla \log \pi(\ba)]$ and $\sigma_-^2 = Var_{\ba \sim \pit^-(\ba)}[\nabla \log \pi(\ba)]$ are the variance of the gradient of the log likelihood inside and outside the buffer. 
If we don't use a baseline, the the optimal sampling strategy is to only sample from $\buf$. 
The variance of the gradient estimation is 
\begin{equation}
\Var[\nabla \objer] \approx \pit(\buf)^2 \sigma_+^2
\end{equation}

If we  use a baseline  $b = \pit(\mathcal{B})$ and apply the optimal sampling allocation  (Section~\ref{apdx:opt-alloc}), then  the variance  of the gradient estimation is
\begin{equation}
\Var[\nabla \objer]_{b} \approx \pit(\mathcal{B})^2(1 - \pit(\mathcal{B}))^2 (\sigma_+^2 + \sigma_-^2)  \label{eq:var_mapob}
\end{equation} 
% (See the Appendix for more details of the proof.)

%In the Appendix, 
We can prove that both of these estimators reduce the variance for the gradient estimation. 
To compare the two, we can see that the ratio of the variance with and without baseline is 
\begin{equation}
\frac{\Var[\nabla \objer]_b}{\Var[\nabla \objer]} = (1 +  \frac{\sigma_-^2}{\sigma_+^2})(1 - \pit(\mathcal{B}))^2  \label{eq:var_mapo}
\end{equation}
%in which $\rho = \frac{\sigma_-^2}{\sigma_+^2}$ is the ratio of the variance of the log likelihood outside and inside the buffer. 
So using baseline provides lower variance when $\pit(\mathcal{B}) \approx 1.0$, which roughly corresponds to the later stage of training, and when $\sigma_-$ is not much larger than $\sigma_+$; it is better to not use baselines when $\pit(\mathcal{B})$ is not close to 1.0 or when $\sigma_-$ is much larger than $\sigma_+$.

\subsection{Optimal Sample Allocation}
\label{apdx:opt-alloc}
%Let's assume that $\buf^+$ and $\buf^-$ represent all the rewarded examples and non-rewarded samples. 
Given that we want to apply stratified sampling to estimate the gradient of \REINFORCE with baseline~\ref{eq:pg},
here we show that the optimal sampling strategy 
%in this case 
is to allocate the same number of samples to $\buf$ and $\A - \buf$. 

Assume that the gradient of log likelihood has similar variance on  $\buf$ and $\A - \buf$:
\begin{equation}
\Var_{\pit^{+}(\ba)}[\nabla \log \pi_{\theta}(\ba)] \approx  \Var_{\pit^{-}(\ba)}[\nabla \log \pi_{\theta}(\ba)] = \sigma^{2}
\end{equation}
%\jb{this is the variance of a vector, so a bit unclear what $\sigma$ is.}
The variance of gradient estimation on $\buf$ and $\A - \buf$ are:
\begin{equation}
\label{eqn:var-strata}
\begin{split}
\Var_{\pit^{+}(\ba)}[(1 - \pit(\buf) )  \nabla \log \pit(\ba)] = & (1 - \pit(\buf) )^2 * \sigma^2 \\
\Var_{\pit^{-}(\ba)}[(- \pit(\buf) )  \nabla \log \pit(\ba)] = & \pit(\buf)^2 * \sigma^2
\end{split}
\end{equation}
%\jb{Is this true because the first term is a constant if it is computed exactly? so you just square and multiply?}

When performing stratified sampling, the optimal sample allocation is to let the number of samples from a stratum be proportional to its probability mass times the standard deviation $P_i \sigma_i$ 
%\jb{I didn't know that - doesn't that introduce bias?}. 
In other words, the more probability mass and the more variance a stratum has, the more samples we should draw from a stratum. So the ratio of the number of samples allocated to each stratum under the optimal sample allocation is
\begin{equation}
\frac{k^+}{k^-} = \frac{\pit(\buf) \sqrt{\Var_{\pit^{+}(\ba)}}}{1 - \pit(\buf) \sqrt{\Var_{\pit^{-}(\ba)}}} 
\end{equation}
Using equation \ref{eqn:var-strata}, we can see that 
\begin{equation}
\frac{k^+}{k^-} = 1
\end{equation}
So the optimal strategy is to allocate the same number of samples to  $\buf$ and $\A - \buf$.
\section{Distributed Actor-Learner Architecture}
\label{apdx:actor-learner}
%\begin{wrapfigure}{r}{2.9in}
\begin{figure}[h!]
  \centering
  %\vspace*{-.1cm}
  %\vspace{-1in}
  \includegraphics[width=4.5in]{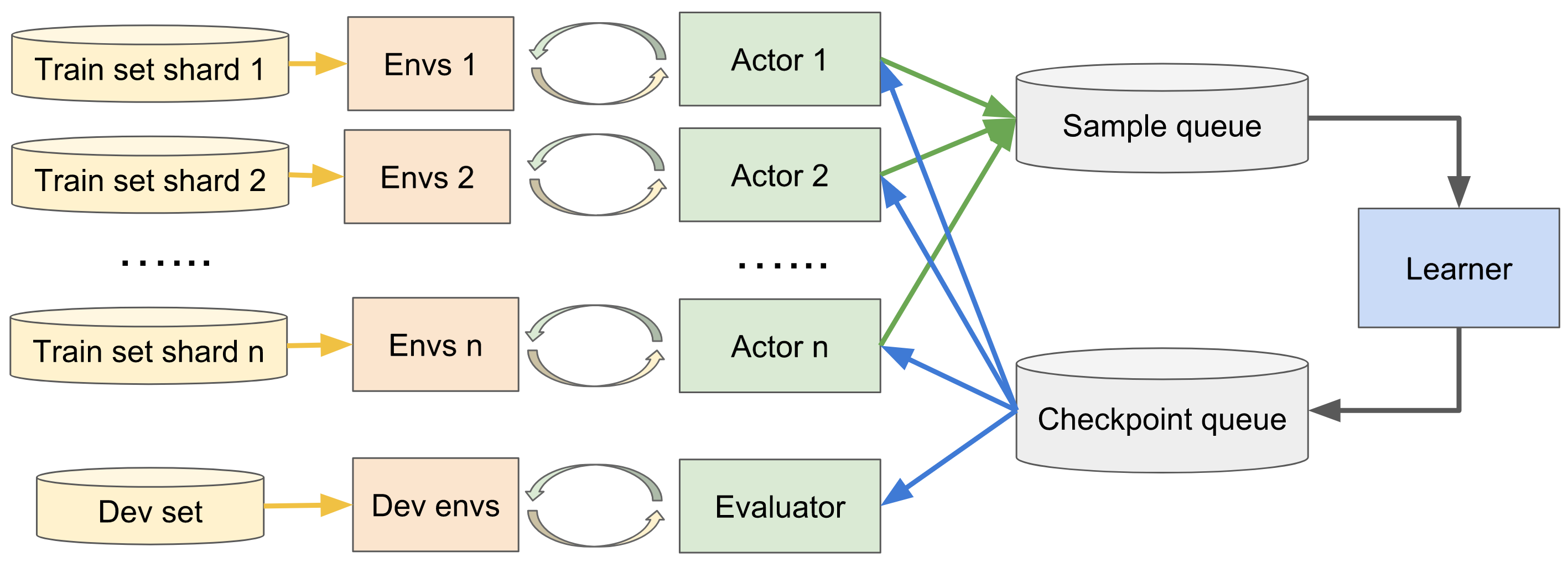}
  %\fbox{\rule[-.5cm]{0cm}{4cm} \rule[-.5cm]{4cm}{0cm}}
%   \vspace{-0.2in}
  \caption{Distributed actor-learner architecture.}
  \label{fig:distributed}
  \vspace{-0.1in}
\end{figure}

Using 30 CPUs, each running one actor, and 2 GPUs, one for training and one for evaluating on dev set, the experiment finishes in about 3 hours on WikiTableQuestions and about 7 hours on WikiSQL. 

\section{Pruning Rules for Random Exploration on WikiTableQuestions}
\label{apdx:prune}

The pruning rules are inspired by the grammar~\cite{zhang2017macro}. It can be seen as trigger words or POS tags for a subset of the functions. For the functions included, they are only allowed when at least one of the trigger words / tags appears in the sentence. For the other functions that are not included, there isn't any constraints. Also note that these rules are only used during random exploration. During training and evaluation, the rules are not applied. 

\begin{table}[!htbp]
%\setlength{\tabcolsep}{2pt}
%\small
\centering 
\begin{tabular}{ll}
\toprule
\bf{Function} & \bf{Triggers} \\
\midrule 
\textbf{count} & how, many, total, number \\
\midrule 
\textbf{filter$_{!in}$} & not, other, besides \\
\midrule 
\textbf{first} & first, top \\
\midrule 
\textbf{last} & last, bottom \\
\midrule 
\textbf{argmin} & JJR, JJS, RBR, RBS, top, first, bottom, last \\
\midrule 
\textbf{argmax} & JJR, JJS, RBR, RBS, top, first, bottom, last \\
\midrule 
\textbf{sum} & all, combine, total \\
\midrule 
\textbf{average} & average \\
\midrule 
\textbf{maximum} & JJR, JJS, RBR, RBS \\
\midrule 
\textbf{minimum} & JJR, JJS, RBR, RBS \\
\midrule 
\textbf{mode} & most \\
\midrule 
\textbf{previous} & next, previous, after, before, above, below \\
\midrule 
\textbf{next} & next, previous, after, before, above, below \\
\midrule
\textbf{same} & same \\
\midrule 
\textbf{diff} & difference, more, than \\
\midrule 
\textbf{filter$_{\geq}$} & RBR, JJR, more, than, least, above, after \\
\midrule 
\textbf{filter$_{\leq}$} & RBR, JJR, less, than, most, below, before, under \\
\midrule 
\textbf{filter$_>$} & RBR, JJR, more, than, least, above, after \\
\midrule 
\textbf{filter$_<$}  & RBR, JJR, less, than, most, below, before, under \\
\bottomrule
\end{tabular}
\label{tab-prune}
\setlength{\abovecaptionskip}{10pt}
\caption{Pruning rules used during random exploration on WikiTableQuestions.}
\vspace{-0.2in}
\end{table}

%\end{wrapfigure}

% \begin{wrapfigure}{r}{2.9in}
% %\begin{figure}
%   \centering
%   \vspace*{-.1cm}
%   \vspace{-0.1in}
%   \includegraphics[width=2.9in]{distributed_rl.png}
%   %\fbox{\rule[-.5cm]{0cm}{4cm} \rule[-.5cm]{4cm}{0cm}}
%   \vspace{-0.2in}
%   \caption{Distributed actor-learner architecture.}
%   \label{fig:distributed}
%   \vspace{-0.1in}
% %\end{figure}
% \end{wrapfigure}
\newpage

% \begin{tabular}{ l ccc}
% \hline
% Type & Question \& Program  \\ 
% \hline
% Superlative & \shortstack{nt-13901: the most points were scored by which player? \\
% (argmax rows r.points-number) 
% (hop v0 r.player-string)
% } \\
% \hline
% same & \shortstack{nt-601: which other ship was launched in the same year as the wave victor? \\
% (filter rows ['wave victor'] r.name-string) 
% (same v0 r.launched-num2)
% (hop v1 r.name-string)} \\
% \hline
% difference & \shortstack{nt-457: what is the difference in child population between koraput and puri? \\
% (filter rows ['koraput'] r.district-string) 
% (filter rows ['puri'] r.district-string) 
% (diff v1 v0 r.child\_population\_0\_6\_years-number)
% } \\
% \hline
% `before or after' & \shortstack{which other ship was launched in the same year as the wave victor? \\
% (filter rows ['wave victor'] r.name-string) 
% (same v0 r.launched-num2) 
% (hop v1 r.name-string)
% } \\
% \hline 
% 

% \caption[Results on the dev]{Results on 
% %Dev and Test sets of
%   WikiTableQuestions dataset. 
%   E.S. is the number of %models used in deep model 
%   ensembles (if applicable).
%   %\protect\footnotemark 
% }
% \label{tab-wtq}

\end{document}